%% file: neurips_2021.tex
\documentclass{article}

\makeatletter
\def\@fnsymbol#1{\ensuremath{\ifcase#1\or \dagger\or \ddagger\or
   \mathsection\or \mathparagraph\or \|\or **\or \dagger\dagger
   \or \ddagger\ddagger \else\@ctrerr\fi}}
   \makeatother

\usepackage[final]{neurips_2021}

\usepackage[utf8]{inputenc} 
\usepackage[T1]{fontenc}    
\usepackage{hyperref}       
\usepackage{url}            
\usepackage{booktabs}       
\usepackage{amsfonts}       
\usepackage{nicefrac}       
\usepackage{microtype}      
\usepackage{xcolor}         
\usepackage{wrapfig}
\usepackage{graphicx}
\usepackage{graphics}
\usepackage{xcolor}
\usepackage[toc,page]{appendix}
\usepackage{subfig}
\usepackage{listings}

\usepackage[frozencache=true,cachedir=.]{minted}

\usepackage{amsmath}
\usepackage{amssymb}
\usepackage{cleveref}
\usepackage{enumitem}
\usepackage{algorithm,algpseudocode}

\newcommand{\etal}{{et al}.\@ }

\makeatletter
\newcommand{\printfnsymbol}[1]{%
  \textsuperscript{\@fnsymbol{#1}}%
}
\makeatother

\title{Compressive Visual Representations}

\author{%
  Kuang-Huei Lee\thanks{Main contributors} \\
  Google Research \\
  \texttt{leekh@google.com} \\
  \And
  Anurag Arnab\printfnsymbol{1} \\
  Google Research \\
  \texttt{aarnab@google.com} \\
  \AND
  Sergio Guadarrama \\
  Google Research \\
  \texttt{sguada@google.com} \\
  \And
  John Canny \\
  Google Research \\
  \texttt{canny@google.com} \\
  \And
  Ian Fischer\printfnsymbol{1} \\
  Google Research \\
  \texttt{iansf@google.com} \\
}

\begin{document}

\maketitle
\setcounter{footnote}{0} 

\input{text/abstract.tex}
\input{text/introduction.tex}
\input{text/methods.tex}
\input{text/experiments.tex}
\input{text/related_work.tex}
\input{text/conclusion.tex}




\begin{ack}
We thank Toby Boyd for his help in making our implementation open source.

John Canny is also affiliated with the University of California, Berkeley.
\end{ack}

\bibliographystyle{plain}
\bibliography{ref}

\newpage

\begin{appendices}

\input{text/appendix.tex}

\end{appendices}
\end{document}

%% file: text/abstract.tex
\begin{abstract}
Learning effective visual representations that generalize well without human supervision is a fundamental problem in order to apply Machine Learning to a wide variety of tasks.
Recently, two families of self-supervised methods, contrastive learning and latent bootstrapping, exemplified by SimCLR and BYOL respectively, have made significant progress.
In this work, we hypothesize that adding explicit information compression to these algorithms yields better and more robust representations.
We verify this by developing SimCLR and BYOL formulations compatible with the Conditional Entropy Bottleneck (CEB) objective, allowing us to both measure and control the amount of compression in the learned representation, and observe their impact on downstream tasks.
Furthermore, we explore the relationship between Lipschitz continuity and compression, showing a tractable lower bound on the Lipschitz constant of the encoders we learn.
As Lipschitz continuity is closely related to robustness, this provides a new explanation for why compressed models are more robust.
Our experiments confirm that adding compression to SimCLR and BYOL significantly improves linear evaluation accuracies and model robustness across a wide range of domain shifts.
In particular, the compressed version of BYOL achieves 76.0\% Top-1 linear evaluation accuracy on ImageNet with ResNet-50, and 78.8\% with ResNet-50 2x.\footnote{Code available at \url{https://github.com/google-research/compressive-visual-representations}}
\end{abstract}

%% file: text/introduction.tex
\section{Introduction}
\label{sec:intro}

Individuals develop mental representations of the surrounding world that generalize over different views of a \textit{shared context}.
For instance, a shared context could be the identity of an object, as it does not change when viewed from different perspectives or lighting conditions.
This ability to represent views by distilling information about the \textit{shared context} has motivated a rich body of self-supervised learning work \cite{ oord2018representation, bachman2019learning, chen2020simple, grill2020bootstrap, he2020momentum, lee2020predictive}.
For a concrete example, we could consider an image from the ImageNet training set \cite{russakovsky2015imagenet} as a shared context, and generate different views by repeatedly applying different data augmentations.
Finding stable representations of a shared context corresponds to learning a minimal high-level description since not all information is relevant or persistent.
This explicit requirement of learning a concise representation leads us to prefer objectives that are \textit{compressive} and only retain the relevant information.

Recent contrastive approaches to self-supervised visual representation learning aim to learn representations that maximally capture the mutual information between two transformed views of an image~\cite{oord2018representation,bachman2019learning, chen2020simple,he2020momentum,hjelm2018learning}.
The primary idea of these approaches is that this mutual information corresponds to a general shared context that is invariant to various transformations of the input, and it is assumed that such invariant features will be effective for various downstream higher-level tasks.
However, although existing contrastive approaches maximize mutual information between augmented views of the same input, they do not necessarily compress away the irrelevant information from these views~\cite{chen2020simple, he2020momentum}.
As shown in~\cite{fischer2020conditional,fischer2020ceb}, retaining irrelevant information often leads to less stable representations and to failures in robustness and generalization, hampering the efficacy of the learned representations.
An alternative state-of-the-art self-supervised learning approach is BYOL~\cite{grill2020bootstrap}, which uses a slow-moving average network to learn consistent, view-invariant representations of the inputs.
However, it also does not explicitly capture relevant compression in its objective.

In this work, we modify SimCLR~\cite{chen2020simple}, a state-of-the-art contrastive representation method, by adding information compression using the Conditional Entropy Bottleneck (CEB)~\cite{fischer2020ceb}.
Similarly, we show how BYOL~\cite{grill2020bootstrap} representations can also be compressed using CEB.
By using CEB we are able to measure and control the amount of information compression in the learned representation~\cite{fischer2020conditional}, and observe its impact on downstream tasks.
We empirically demonstrate that our compressive variants of SimCLR and BYOL, which we name C-SimCLR and C-BYOL, significantly improve accuracy and robustness to domain shifts across a number of scenarios.
Our primary contributions are:
\begin{itemize}[leftmargin=1em]
\item Reformulations of SimCLR and BYOL such that they are compatible with information-theoretic compression using the Conditional Entropy Bottleneck \cite{fischer2020conditional}.
\item An exploration of the relationship between Lipschitz continuity, SimCLR, and CEB compression, as well as a simple, tractable lower bound on the Lipschitz constant.
This provides an alternative explanation, in addition to the information-theoretic view \cite{fischer2020conditional, fischer2020ceb,achille2017emergence, achille2018information}, for why CEB compression improves SimCLR model robustness.
\item Extensive experiments supporting our hypothesis that adding compression to the state-of-the-art self-supervised representation methods like SimCLR and BYOL can significantly improve their performance and robustness to domain shifts across multiple datasets.
In particular, linear evaluation accuracies of C-BYOL are even competitive with the supervised baselines considered by SimCLR~\cite{chen2020simple} and BYOL~\cite{grill2020bootstrap}.
C-BYOL reaches 76.0\% and 78.8\% with ResNet-50 and ResNet-50 2x respectively, whereas the corresponding supervised baselines are 76.5\% and 77.8\% respectively.
\end{itemize}

%% file: text/methods.tex
\section{Methods}
\label{sec:methods}

In this section, we describe the components that allow us to make distributional, compressible versions of SimCLR and BYOL.
This involves switching to the Conditional Entropy Bottleneck (CEB) objective, noting that the von Mises-Fisher distribution is the exponential family distribution that corresponds to the cosine similarity loss function used by SimCLR and BYOL, and carefully identifying the random variables and the variational distributions needed for CEB in SimCLR and BYOL.
We also note that SimCLR and CEB together encourage learning models with a smaller Lipschitz constant, although they do not explicitly enforce that the Lipschitz constant be small.

\subsection{The Conditional Entropy Bottleneck}
\label{sec:ceb}

In order to test our hypothesis that compression can improve visual representation quality, we need to be able to measure and control the amount of compression in our visual representations.
To achieve this, we use the Conditional Entropy Bottleneck (CEB)~\citep{fischer2020conditional}, an objective function in the Information Bottleneck (IB)~\citep{tishby2000information} family.

Given an observation $X$, a target $Y$, and a learned representation $Z$ of $X$, CEB can be written as:
\begin{align}
CEB &\equiv \min_Z \beta I(X;Z|Y) - I(Y;Z) \\
&= \min_Z \beta (H(Z) - H(Z|X) - H(Z) + H(Z|Y)) - H(Y) + H(Y|Z) \\
&= \min_Z \beta(-H(Z|X) + H(Z|Y)) + H(Y|Z) \label{eq:ceb}
\end{align}
where $H(\cdot)$ and $H(\cdot|\cdot)$ denote entropy and conditional entropy respectively. 
We can drop the $H(Y)$ term because it is constant with respect to $Z$.
$I(Y;Z)$ is the useful information relevant to the task, or the prediction target $Y$.
$I(X;Z|Y)$ is the \emph{residual information} $Z$ captures about $X$ when we already know $Y$, which we aim to minimize.
Compression strength increases as $\beta$ increases.

We define $e(z|x)$ as the true encoder distribution, where $z$ is sampled from; $b(z|y)$, a variational approximation conditioned on $y$; $d(y|z)$, the decoder distribution (also a variational approximation) which predicts $y$ conditioned on $z$.
As shown in~\cite{fischer2020conditional}, CEB can be variationally upper-bounded:
\begin{align}
vCEB &\equiv \min_{e(z|x),b(z|y),d(y|z)} \mathbb{E}_{x,y \sim p(x,y), z \sim e(z|x)} \beta(\log e(z|x) - \log b(z|y)) - \log d(y|z)
\label{eq:vceb}
\end{align}
There is no requirement that all three distributions have learned parameters.
At one limit, a model's parameters can be restricted to any one of the three distributions; at the other limit, all three distributions could have learned parameters.
If $e(\cdot)$ has learned parameters, its distributional form may be restricted, as we must be able to take gradients through the $z$ samples.\footnote{%
  For example, $e(z|x)$ could not generally be a mixture distribution, as sampling the mixture distribution has a discrete component, and we cannot easily take gradients through discrete samples.
}
The only requirement on the $b(\cdot)$ and $d(\cdot)$ distributions is that we be able to take gradients through their log probability functions.

\paragraph{InfoNCE.}
As shown in~\cite{fischer2020conditional}, besides parameterizing $d(y|z)$, it is possible to reuse $b(z|y)$ to make a variational bound on the $H(Y|Z)$ term.
As $I(Y;Z) = H(Y) - H(Y|Z)$ and $H(Y)$ is a constant with respect to $Z$:
\begin{align}
\label{eq:infonce}
H(Y|Z) \leq E_{x,y \sim p(x,y), z \sim e(z|x)} \log \frac{b(z|y)}{\sum_{k=1}^K b(z|y_k)}
\end{align}
where $K$ is the number of examples in a minibatch.
Eq.~\eqref{eq:infonce} is also known as the contrastive \textit{InfoNCE} bound~\cite{oord2018representation,vmibounds}.
The inner term,
\begin{align}
    d(y|z) \equiv \frac{b(z|y)}{\sum_{k=1}^K b(z|y_k)},
    \label{eq:catgen}
\end{align}
is a valid variational approximation of the true but unknown $p(y|z)$.
Fischer~\cite{fischer2020conditional} calls Eq.~\eqref{eq:catgen} the \textit{CatGen} decoder because it is a categorical distribution over the minibatch that approximates the generative decoder distribution.

\subsection{C-SimCLR: Compressed SimCLR}
\label{sec:simclr}

The InfoNCE bound \cite{oord2018representation} enables many contrastive visual representation methods to use it to capture shared context between different views of an image as a self-supervised objective \cite{chen2020simple,chen2020big,he2020momentum,chen2020exploring,hjelm2018learning}.
In this work, we show how to compress the SimCLR \cite{chen2020simple} model, but the method we discuss is generally applicable to other InfoNCE-based models.

SimCLR applies randomized augmentations to an image to create two different views, $x$ and $y$ (which we also refer to as $x'$), and encodes both of them with a shared encoder, producing representations $r_x$ and $r_y$.
Both $r_x$ and $r_y$ are $l_2$-normalized.
The SimCLR version of the InfoNCE objective is:
\begin{align}
\label{eq:simclr_nce}
L_{NCE}(r_x, r_y) = -\log \frac{e^{\frac{1}{\tau}r_y^Tr_x}}{\sum_{k=1}^Ke^{\frac{1}{\tau}r_{y_k}^Tr_x}}
\end{align}
where $\tau$ is a temperature term and $K$ is the number of views in a minibatch.
SimCLR further makes its InfoNCE objective \textit{bidirectional}, such that the final objective becomes
\begin{align}
L_{NCE}(r_x, r_y) + L_{NCE}(r_y, r_x) 
= -\log \frac{e^{\frac{1}{\tau}r_y^Tr_x}}{\sum_{k=1}^Ke^{\frac{1}{\tau}r_{y_k}^Tr_x}} - \log \frac{e^{\frac{1}{\tau}r_x^Tr_y}}{\sum_{k=1}^Ke^{\frac{1}{\tau}r_{x_k}^Tr_y}}
\end{align}
We can observe the following: 
$\exp(\frac{1}{\tau} r_y^Tr_x)$ in Eq.~\eqref{eq:simclr_nce} corresponds to the unnormalized $b(z|y)$ in Eq.~\eqref{eq:infonce}.
$e(\cdot|x)$ generates $z=r_x$, whilst
$r_y$ and $r_{y_k}$ are distribution parameters of $b(\cdot|y)$ and $b(\cdot|y_k)$ respectively.
$e(\cdot|x)$ and $b(\cdot|y)$ share model parameters.

\paragraph{von Mises-Fisher Distributional Representations.}
\label{sec:vmf}

The cosine-similarity-based loss (Eq.~\eqref{eq:simclr_nce}) is commonly used in contrastive learning and can be connected to choosing the von Mises-Fisher (vMF) distribution for $e(\cdot|x)$ and $b(\cdot|y)$ \cite{hasnat2017mises,wang2020understanding}.
vMF is a distribution on the $(n-1)$-dimensional hyper-sphere. 
The probability density function is given by $f_n(z, \mu, \kappa) = C_n(\kappa)e^{\kappa \mu^Tz}$, where $\mu$ and $\kappa$ are denoted as the mean direction and concentration parameter respectively.
We assume $\kappa$ is a constant.
The normalization term $C_n(\kappa)$ is a function of $\kappa$ and equal to $\frac{\kappa^{n/2-1}}{(2\pi)^{n/2} I_{n/2-1}(\kappa)}$, where $I_v$ denotes the modified Bessel function of the first kind at order $v$.

By setting the mean direction $\mu$ to $r_y$, concentration $\kappa_b$ of $b(\cdot|y)$ to $1/\tau$, and $r_x$ to $z$, we can connect the SimCLR objective (Eq.~\eqref{eq:simclr_nce}) to the distributional form of InfoNCE (Eq.~\eqref{eq:infonce})
\begin{align}
\frac{e^{\frac{1}{\tau}r_y^Tr_x}}{\sum_{k=1}^Ke^{\frac{1}{\tau}r_{y_k}^Tr_x}}
= \frac{C_n(\kappa_b)e^{\kappa_b r_y^Tr_x}}{\sum_{k=1}^K C_n(\kappa_b) e^{\kappa_b r_{y_k}^Tr_x}}
= \frac{f_n(r_x, r_y, \kappa_b)}{\sum_{k=1}^K f_n(r_x, r_{y_k}, \kappa_b)}
= \frac{b(r_x|y)}{\sum_{k=1}^K b(r_x|y_k)}
\end{align}
$z=r_x$ is a deterministic unit-length vector, so we can view $e(\cdot|x)$ as a spherical delta distribution, which is equivalent to a vMF with $r_x$ as the mean direction and $\kappa_e \rightarrow \infty$.
We can further extend the forward encoder to have non-infinite $\kappa_e$, which results in a stochastic $z$.
These allow us to have SimCLR in a distributional form with explicit distributions $e(\cdot|x)$ and $b(\cdot|y)$ and satisfy the requirements of CEB discussed in Sec.~\ref{sec:ceb}.

\begin{figure}
    \centering
    \includegraphics[keepaspectratio, width=0.8\textwidth]{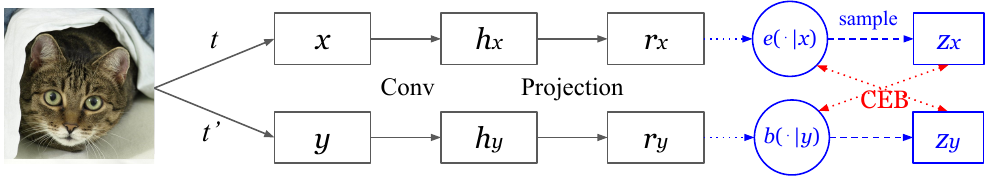}
    \caption{C-SimCLR explicitly defines encoder distributions $e(\cdot|x)$ and $b(\cdot|y)$ where $x$ and $y$ are two augmented views of an image. $y$ is also referred as $x'$. The upper and lower encoder outputs are used to specify mean directions of $e$ and $b$, and the two encoders share parameters. $r_x, r_y$ are $l_2$-normalized. Our modifications to SimCLR are highlighted in blue. No new parameters are added.}
    \label{fig:ceb_simclr}
\end{figure}

\paragraph{Compressing SimCLR with Bidirectional CEB.}
\Cref{fig:ceb_simclr} illustrates the Compressed SimCLR (C-SimCLR) model.
The model learns a compressed representation of an view $X$ that only preserves information relevant to predicting a different view $Y$ by switching to CEB.
As can be seen in Eq.~\eqref{eq:ceb}, the CEB objective treats $X$ and $Y$ asymmetrically.
However, as shown in~\cite{fischer2020conditional}, it is possible to learn a single representation $Z$ of both $X$ and $Y$ by having the forward and backward encoders act as variational approximations of each other:
\begin{align}
CEB_{\text{bidir}} &\equiv \min_Z \beta_X I(X;Z|Y) - I(Y;Z) + \beta_Y I(Y;Z|X) - I(X;Z) \\
&\equiv \min_Z \beta_X(-H(Z|X) + H(Z|Y)) + H(Y|Z) \\
&~~~~~~~~~~~~ + \beta_Y(-H(Z|Y) + H(Z|X)) + H(X|Z) \nonumber \\
&\leq \min_{e(\cdot|\cdot),b(\cdot|\cdot),c(\cdot|\cdot),d(\cdot|\cdot)} \mathbb{E}_{x,y \sim p(x,y)} \Big[ \label{eq:ceb_bidir} \\
&~~~~~~~~~~~~~~~~~~ \mathbb{E}_{z_x \sim e(z_x|x)} \big[ \beta_X(\log e(z_x|x) - \log b(z_x|y)) - \log d(y|z_x) \big] \nonumber \\
&~~~~~~~~~~~~~~~~~~ + \mathbb{E}_{z_y \sim e(z_y|y)} \big[ \beta_Y(\log(e(z_y|y) - \log(b(z_y|x)) - \log c(x|z_y) \big] \Big] \nonumber
\end{align}
where $d(\cdot|\cdot)$ and $c(\cdot|\cdot)$ are the InfoNCE variational distributions of $b(\cdot|\cdot)$ and $e(\cdot|\cdot)$ respectively.
$e$ and $b$ use the same encoder to parameterize mean direction in SimCLR setting.
Since SimCLR is trained with a bidirectional InfoNCE objective, Eq.~\eqref{eq:ceb_bidir} gives an easy way to compress its learned representation.
As in SimCLR, the deterministic $h_x$ (in Fig.~\ref{fig:ceb_simclr}) is still the representation used on downstream classification tasks.

\subsection{C-BYOL: Compressed BYOL}
\label{sec:byol}

\begin{figure}
    \centering
    \includegraphics[keepaspectratio, width=1.0\textwidth]{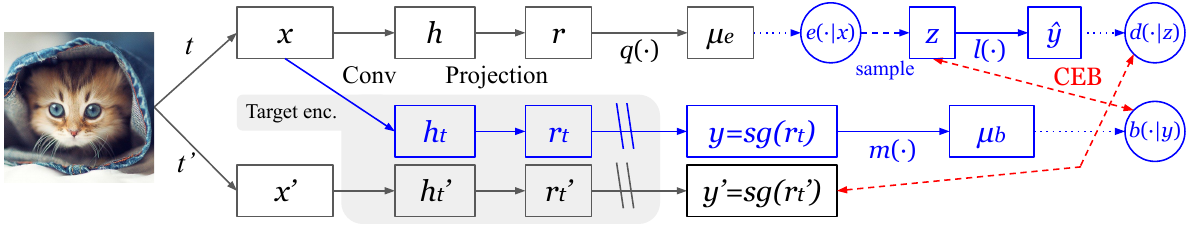}
    \caption{C-BYOL. The upper online encoder path takes an augmented view $x$ as input and produces $e(\cdot|x)$ and $d(\cdot|z)$. The lower two paths use the same target encoder (shaded), which is a moving average of the online encoder (Conv + Projection). The target encoder maps $x$ and another view $x'$ to $r_t$ and $r_t'$. $sg(r_t)$ ($sg$: stop gradients) is our target $y$. $y$ leads to $b(\cdot|y)$. $sg(r_t')$ is our perturbed target $y'$. $r_t, r_t', \mu_e, \mu_b, \hat{y}$ are $l_2$-normalized. These yield the components required by CEB. We highlight changes to BYOL in blue.}
    \label{fig:byol}
\end{figure}

In this section we will describe how to modify BYOL to make it compatible with CEB, as summarized in Fig.~\ref{fig:byol}.
BYOL~\cite{grill2020bootstrap} learns an online encoder that takes $x$, an augmented view of a given image, as input and predicts outputs of a target encoder which encodes $x'$, a different augmented view of the same image.
The target encoder's parameters are updated not by gradients but as an exponential moving average of the online encoder's parameters.
The loss function is simply the mean square error, which is equivalent to the cosine similarity between the online encoder output $\mu_e$ and the target encoder output $y'$ as both $\mu_e$ and $y'$ are $l_2$-normalized:
\begin{align}
    L_{byol} = ||\mu_e - y'||^2_2 = \mu_e^T\mu_e + {y'}^T{y'} - 2 \mu_e^Ty' = 2 - 2 \mu_e^Ty'
\label{eq:byol_loss}
\end{align}
This iterative ``latent bootstrapping'' allows BYOL to learn a view-invariant representation.
In contrast to SimCLR, BYOL does not rely on other samples in a batch and does not optimize the InfoNCE bound.
It is a simple regression task: given input $x$, predict $y'$.
To make BYOL CEB-compatible, we need to identify the random variables $X$, $Y$, $Z$, define encoder distributions $e(z|x)$ and $b(z|y)$, and define the decoder distribution $d(y|z)$ (see \Cref{eq:vceb}). 

We define $e(z|x)$ to be a vMF distribution parameterized by $\mu_e$, and sample $z$ from $e(z|x)$:
\begin{align}
    e(z|x) = C_n(\kappa_e)e^{\kappa_e z^T\mu_e}
\label{eq:byol_ezx}
\end{align}
We use the target encoder to encode $x$ and output $r_t$, an $l_2$-normalized vector. 
We choose $r_t$ to be $y$. 
We then add a 2-layer MLP on top of $y$ and $l_2$-normalize the output, which gives $\mu_b$.
We denote this transformation as $\mu_b=m(y)$ and define $b(z|y)$ to be the following vMF parameterized by $\mu_b$:
\begin{align}
    b(z|y) = C_n(\kappa_b)e^{\kappa_b z^T\mu_b}
\label{eq:byol_bzy}
\end{align}
For $d(y|z)$, we add a linear transformation on $z$ with $l_2$-normalization, $\hat{y}=l(z)$, and define a vMF parameterized by $\hat{y}$:
\begin{align}
d(y|z) = C_n(\kappa_d)e^{\kappa_d y^T\hat{y}}
\label{eq:byol_dyz}
\end{align}
In the deterministic case where $z$ is not sampled, this corresponds to adding a linear layer with $l_2$-normalization on $\mu_e$ which does not change the model capacity and empirical performance.

In principle, we can use any stochastic function of $Z$ to generate $Y$.
In our implementation, we replace the generative decoder $\log d(y|z)$ with $\log d(y'|z)$, where we use the target encoder to encode $x'$ and output $y'$.
Given that $X \rightarrow X'$ is a stochastic transformation and both $X$ and $X'$ go through the same the target encoder function, $Y \rightarrow Y'$ is also a stochastic transformation.
$d(y'|z)$ can be considered as having a stochastic perturbation to $d(y|z)$.
Our $vCEB$ objective becomes
\begin{align}
    L_{cbyol}(x, x') = \beta(\log e(z|x) - \log b(z|y)) - \log d(y'|z).
\end{align}
We empirically observed the best results with this design choice. 
$d(y'|z)$ can be directly connected the standard BYOL regression objective:
When $\kappa_d=2$, $-\log(d(y'|z)) = -\kappa_d y'^T\hat{y} -\log(C_n(\kappa_d))$ is equivalent to Eq.~\eqref{eq:byol_loss} when constants are ignored.

Although it seems that we additionally apply the target encoder to $x$ compared to BYOL, this does not increase the computational cost in practice. 
As in BYOL, the learning objective is applied symmetrically in our implementation: $L_{cbyol}(x, x') + L_{cbyol}(x', x)$.
Therefore, the target encoder has to be applied to both $x$ and $x'$ no matter in BYOL or C-BYOL.
Finally, note that like in BYOL, $h$ (Fig.~\ref{fig:byol}) is the deterministic representation used for downstream tasks.

\subsection{Lipschitz Continuity and Compression}
\label{sec:lipschitz}

Lipschitz continuity provides a way of measuring how smooth a function is.
For some function $f$ and a distance measure $D(f(x_1), f(x_2))$, Lipschitz continuity defines an upper bound on how quickly $f$ can change as $x$ changes:
\begin{align}
L ||\Delta x|| \geq D(f(x), f(x + \Delta x)),
\end{align}
where $L$ is the Lipschitz constant, $\Delta x$ is the vector change in $x$, and $||\Delta x|| > 0$.
If we define $f(x)$ to be our encoder distribution $e(z|x)$ (which is a vMF and always positive), and the distance measure, $D$, to be the absolute difference of the logs of the functions, we get a function of $z$ of Lipschitz value, such that:
\begin{align}
\label{eq:encoder_lipschitz_main}
L(z) \geq \frac{1}{||\Delta x||} |\log e(z|x) - \log e(z|x + \Delta x)|
\end{align}

As detailed in Sec.~\ref{sec:complete_lipschitz}, by taking expectations with respect to $z$, we can obtain a lower bound on the encoder \emph{distribution}'s squared Lipschitz constant:\footnote{%
  Note that by taking an expectation we get a KL divergence, which violates the triangle inequality, even though we started from a valid distance metric.
  Squaring the Lipschitz constant addresses this in the common case where the $\operatorname{KL}$ divergence grows quadratically in $||\Delta x||$, as detailed in Section~\ref{sec:complete_lipschitz}.
}
\begin{align}
\label{eq:lipschitz_log_bound1}
L^2 \geq \frac{1}{||\Delta x||^2} \max \Big( \operatorname{KL}[ e(z|x) || e(z|x + \Delta x) ],\, \operatorname{KL}[ e(z|x + \Delta x) || e(z|x) ] \Big)
\end{align}
To guarantee smoothness of the encoder distribution, we would like to have an upper bound on $L$, rather than a lower bound.
Minimizing a lower bound does not directly yield any optimality guarantees relative to the bounded quantity.
However, in this case, minimizing the symmetric $\operatorname{KL}$ below is \emph{consistent} with learning a smoother encoder function:
\begin{align}
\label{eq:min_symkl1}
\inf_{e(z|\cdot)} \operatorname{KL}[ e(z|x) || e(z|x + \Delta x) ] + \operatorname{KL}[ e(z|x + \Delta x) || e(z|x) ]
\end{align}
By \emph{consistent}, we mean that, if we could minimize this symmetric KL at every pair $(x, x + \Delta x)$ in the input domain, we would have smoothed the model.
In practice, for high-dimensional input domains, that is not possible, but minimizing Eq.~\eqref{eq:min_symkl1} at a subset of the input domain still improves the model's smoothness, at least at that subset.

The minimization in Eq.~\eqref{eq:min_symkl1} corresponds almost exactly to the CEB compression term in the bidirectional SimCLR models.
We define $y = x + \Delta x$.
At samples of the augmented observed variables, $X, Y$, the C-SimCLR models minimize upper bounds on the two residual informations:
\begin{align}
\label{eq:bidir_residual1}
I(X;Z|Y) + I(Y;Z|X) \leq \mathbb{E}_{x,y \sim p(x,y)} \operatorname{KL}[ e(z|x) || e(z|y) ] + \operatorname{KL}[ e(z|y) || e(z|x) ]
\end{align}
The only caveat to this is that we use $b(z|y)$ instead of $e(z|y)$ in C-SimCLR. 
$b$ and $e$ share weights but have different $\kappa$ values in their vMF distributions.
However, these are hyperparameters, so they are not part of the trained model parameters.
They simply change the minimum attainable $\operatorname{KL}$s in Eq.~\eqref{eq:bidir_residual1}, thereby adjusting the minimum achievable Lipschitz constant for the models (see Sec.~\ref{sec:complete_lipschitz}).

Directly minimizing \Cref{eq:lipschitz_log_bound1} would require normalizing the symmetric $\operatorname{KL}$ per-example by $||\Delta x||^2$.
The symmetric CEB loss does not do this.
However, the residual information terms in \Cref{eq:bidir_residual1} are multiplied by a hyperparameter $\beta \leq 1$.
Under a simplifying assumption that the $||\Delta x||$ values generated by the sampling procedure are typically of similar magnitude, we can extract the average $\frac{1}{||\Delta x||^2}$ into the hyperparameter $\beta$.
We note that in practice using per-example values of $||\Delta x||^2$ would encourage the optimization process to smooth the model more strongly at observed $(x, \Delta x)$ pairs where it is least smooth, but we leave such experiments to future work.

Due to Eq.~\eqref{eq:bidir_residual1}, we should expect that the C-SimCLR models are locally more smooth around the observed data points.
We reiterate, though, that this is not a proof of increased global Lipschitz smoothness, as we are minimizing a lower bound on the Lipschitz constant, rather than minimizing an upper bound.
It is still theoretically possible to learn highly non-smooth functions using CEB in this manner.
It would be surprising, however, if the C-SimCLR were somehow \emph{less} smooth than the corresponding SimCLR models.

The Lipschitz continuity property is closely related to model robustness to perturbations \cite{bruna2013invariant}, including robustness to adversarial examples \cite{weng2018evaluating, fazlyab2019efficient, yang2020closer}.
Therefore, we would expect to see that the C-SimCLR models are more robust than SimCLR models on common robustness benchmarks.
It is more difficult to make the same theoretical argument for the C-BYOL models, as they do not use exactly the same encoder for both $x$ and $y$.
Thus, the equivalent conditional information terms from Eq.~\eqref{eq:bidir_residual1} are not directly minimizing a lower bound on the Lipschitz constant of the encoder.
Nevertheless, we empirically explore the impact of CEB on both SimCLR and BYOL models next in Sec.~\ref{sec:experiments}.

%% file: text/experiments.tex
\section{Experimental Evaluation}
\label{sec:experiments}

We first describe our experimental set-up in Sec.~\ref{sec:experiments_setup}, before evaluating the image representations learned by our self-supervised models in linear evaluation settings in Sec.~\ref{sec:experiments_imagenet}.
We then analyse the robustness and generalization of our self-supervised representations by evaluating model accuracy across a wide range of domain and distributional shifts in Sec.~\ref{sec:experiments_robustness}.
Finally, we analyse the effect of compression strength in Sec.~\ref{sec:experiments_ablations}.
Additional experiments and ablations can be found in the Appendix.

\subsection{Experimental Set-up}
\label{sec:experiments_setup}

\paragraph{Implementation details.}
Our implementation of SimCLR, BYOL, and their compressed versions is based off of the public implementation of SimCLR~\cite{chen2020simple}.
Our implementation consistently reproduces BYOL results from \cite{grill2020bootstrap} and outperforms the original SimCLR, as detailed in Sec.~\ref{sec:impl_detail}.

We use the same set of image augmentations as in BYOL \cite{grill2020bootstrap} for both BYOL and SimCLR, and also use BYOL's (4096, 256) two-layer projection head for both methods.
We follow SimCLR and BYOL to use the LARS optimizer~\cite{you2017large} with a cosine decay learning rate schedule \cite{loshchilov2016sgdr} over 1000 epochs with a warm-up period, as detailed in Sec.~\ref{sec:detail_training}.
For ablation experiments we train for 300 epochs instead.
As in SimCLR and BYOL, we use batch size of 4096 split over 64 Cloud TPU v3 cores.
Except for ablation studies of compression strength, $\beta$ is set to $1.0$ for both C-SimCLR and C-BYOL.
We follow SimCLR and BYOL in their hyperparameter choices unless otherwise stated, and provide exhaustive details in Sec.~\ref{sec:impl_detail}.
Pseudocode can be found in Sec.~\ref{sec:pseudocode}.

\paragraph{Evaluation protocol.}
We assess the performance of representations pretrained on the ImageNet training set \cite{russakovsky2015imagenet} without using any labels.
Then we train a linear classifier on different labeled datasets on top of the frozen representation.
The final performance metric is the accuracy of these classifiers.
As our approach builds on SimCLR~\cite{chen2020big} and BYOL~\cite{grill2020bootstrap}, we follow the same evaluation protocols.
Further details are in Sec.~\ref{sec:detail_eval}.

\subsection{Linear Evaluation of Self-supervised Representations}
\label{sec:experiments_imagenet}

\paragraph{Linear evaluation on ImageNet.}
We first evaluate the representations learned by our models by training a linear classifier on top of frozen features on the ImageNet training set, following standard practice~\cite{chen2020simple, grill2020bootstrap, kolesnikov2019revisiting, kornblith2019better}.
As shown in \Cref{tab:linear_eval_r50}, our compressed objectives provide strong improvements to state-of-the-art SimCLR~\cite{chen2020simple} and BYOL~\cite{grill2020bootstrap} models across different ResNet architectures \cite{he2016deep} of varying widths (and thus number of parameters) \cite{zagoruyko2016wide}.
Our reproduction of the SimCLR baseline (70.7\% top-1 accuracy) outperforms that of the original paper (69.3\%).
Our implementation of BYOL, which obtains a mean Top-1 accuracy of 74.2\% (averaged over three trials) matches that of~\cite{grill2020bootstrap} within a standard deviation.

Current self-supervised methods benefit from longer training schedules~\cite{chen2020simple, chen2020big, grill2020bootstrap, chen2020exploring,he2020momentum}.
\Cref{tab:linear_eval_r50} shows that our improvements remain consistent for both 300 epochs, and the longer 1000 epoch schedule which achieves the best results.
In addition to the Top-1 and Top-5 accuracies, we also compute the Brier score~\cite{brier1950verification} which measures model calibration.
Similar to the predictive accuracy, we observe that our compressed models obtain consistent improvements.

\input{tables/imagenet_linear_eval.tex}

\paragraph{Learning with a few labels on ImageNet.}
\input{tables/imagenet_semi_supervised.tex}

After self-supervised pretraining on ImageNet, we learn a linear classifier on a small subset (1\% or 10\%) of the ImageNet training set, using the class labels this time, following the standard protocol of~\cite{chen2020simple, grill2020bootstrap}.
We expect that with strong feature representations, we should be able to learn an effective classifier with limited training examples.

\Cref{tab:semi} shows that the compressed models once again outperform the SimCLR and BYOL counterparts.
The largest improvements are observed in the low-data regime, where we improve upon the state-of-the-art BYOL by 5.1\% and SimCLR by 1.8\%, when using only 1\% of the ImageNet labels.
Moreover, note that self-supervised representations significantly outperform a fully-supervised ResNet-50 baseline which overfits significantly in this low-data scenario.

\paragraph{Comparison to other methods.}

\Cref{tab:linear_eval_sota} compares C-SimCLR and C-BYOL to other recent self-supervised methods from the literature (in the standard setting of using two augmented views) on ImageNet linear evaluation accuracy.
We present accuracy for models trained for 800 and 1000 epochs, depending on what the original authors reported.
C-BYOL achieves the best results compared to other state-of-the-art methods.
Moreover, we can improve C-BYOL with ResNet-50 even further to 76.0 Top-1 accuracy when we train it for 1500 epochs.

\paragraph{Comparison to supervised baselines.}

As shown in \Cref{tab:linear_eval_sota}, SimCLR and BYOL use supervised baselines of 76.5\% for ResNet-50 and 77.8\% for ResNet-50 2x~\cite{chen2020simple,grill2020bootstrap} respectively.
In comparison, the corresponding compressed BYOL models achieve 76.0\% for ResNet-50 and 78.8\% for ResNet-50 2x, effectively matching or surpassing reasonable supervised baselines.\footnote{
  We note that comparing supervised and self-supervised methods is difficult, as it can only be system-wise.
  Various complementary techniques can be used to further improve evaluation results in both settings.
  For example, the appendix of~\cite{grill2020bootstrap} reports that various techniques improve supervised model accuracies, whilst~\cite{grill2020bootstrap, kolesnikov2019revisiting} report various techniques to improve evaluation accuracy of self-supervised representations.
  We omit these in order to follow the common supervised baselines and standard evaluation protocols used in prior work.
}

The results in \Cref{tab:linear_eval_r50,tab:semi,tab:linear_eval_sota} support our hypothesis that compression of SSL techniques can improve their ability to generalize in a variety of settings.
These results are consistent with theoretical understandings of the relationship between compression and generalization~\cite{shamir2008learning,vera2018role,dubois2020learning}, as are the results in \Cref{tab:ceb_ablation} that show that performance improves with increasingly strong compression (corresponding to higher values of $\beta$), up to some maximum amount of compression, after which performance degrades again.

\subsection{Evaluation of Model Robustness and Generalization}
\label{sec:experiments_robustness}

In this section, we analyse the robustness of our models to various domain shifts.
Concretely, we use the models, with their linear classifier, from the previous experiment, and evaluate them on a suite of robustness benchmarks that have the same label set as the ImageNet dataset.
We use the public robustness benchmark evaluation code of~\cite{djolonga2020robustness, djolonga2020robustnesscode}. 
As a result, we can evaluate our network and report Top-1 accuracy, as shown in \Cref{tab:robustness}, without any modifications to the network.

We consider ``natural adversarial examples'' with ImageNet-A~\cite{hendrycks2019benchmarking} which consists of difficult images which a ResNet-50 classifier failed on.
ImageNet-C~\cite{hendrycks2019benchmarking} adds synthetic corruptions to the ImageNet validation set, 
ImageNet-R~\cite{hendrycks2020many} considers other naturally occuring distribution changes in image style while ObjectNet~\cite{barbu2019objectnet} presents a more difficult test set for ImageNet where the authors control for different parameters such as viewpoint and background.
ImageNet-Vid and YouTube-BB~\cite{shankar2019image} evaluate the robustness of image classifiers to natural perturbations arising in video.
Finally, ImageNet-v2~\cite{recht2019imagenet} is a new validation set for ImageNet where the authors attempted to replicate the original data collection process.
Further details of these robustness benchmarks are in \Cref{sec:robustness_benchmark_details}.

\input{tables/robustness}
\Cref{tab:robustness} shows that SimCLR and BYOL models trained with CEB compression consistently outperform their uncompressed counterparts across all seven robustness benchmarks.
This is what we hypothesized in the SimCLR settings based on the Lipschitz continuity argument in Sec.~\ref{sec:lipschitz} and the appendix.
All models performed poorly on ImageNet-A, but this is not surprising given that ImageNet-A was collected by~\cite{hendrycks2019benchmarking} according to images that a ResNet-50 classifier trained with full supervision on ImageNet misclassified, and we evaluate with ResNet-50 models too.

\subsection{The Effect of Compression Strength}
\label{sec:experiments_ablations}
\input{tables/ceb_beta_ablation}

\Cref{tab:ceb_ablation} studies the effect of the CEB compression term, $\beta$ on linear evaluation accuracy on ImageNet, as well as on the same suite of robustness datasets.
We observe that $\beta = 0$, which corresponds to no explicit compression, but a stochastic representation, already achieves improvements across all datasets.
Further improvements are observed by increasing compression ($\beta$), with $\beta = 1$ obtaining the best results.
But overly strong compression can be harmful.
Large values of $\beta$ correspond to high levels of compression, and can cause training to collapse, which we observed for $\beta = 2$.

%% file: tables/imagenet_linear_eval.tex
\begin{table}
  \caption{ImageNet accuracy of linear classifiers trained on representations learned with SimCLR \cite{chen2020simple} and BYOL \cite{grill2020bootstrap}, with and without CEB compression.
  A lower Brier score corresponds to better model calibration.
  We report mean accuracy and standard deviations over three trials.
  }
  \small
  \centering
  \begin{tabular}{lccc|ccc}
    \toprule
    & \multicolumn{3}{c}{SimCLR} & \multicolumn{3}{c}{BYOL} \\
    Method & Top-1 & Top-5 & Brier & Top-1 & Top-5 & Brier \\
    \midrule
    \multicolumn{5}{l}{\textit{ResNet-50, 300 epochs}} \\
    Uncompressed & 69.1{\scriptsize $\pm0.089$} & 89.1{\scriptsize $\pm0.034$} & 42.1{\scriptsize $\pm1.06$} & 72.8{\scriptsize $\pm0.155$} & 91.0{\scriptsize $\pm0.072$} & 37.3{\scriptsize $\pm0.089$}  \\
    Compressed & \textbf{70.1}{\scriptsize $\pm0.177$} & \textbf{89.6}{\scriptsize $\pm0.099$} & \textbf{41.0}{\scriptsize $\pm0.107$} & \textbf{73.6}{\scriptsize $\pm0.039$} & \textbf{91.5}{\scriptsize $\pm0.080$} & \textbf{36.5}{\scriptsize $\pm0.045$}\\
    \midrule
    \multicolumn{5}{l}{\textit{ResNet-50, 1000 epochs}} \\
    Uncompressed & 70.7{\scriptsize $\pm0.094$} & 90.1{\scriptsize $\pm0.081$} &40.0{\scriptsize $\pm0.123$} & 74.2{\scriptsize $\pm0.139$} & 91.7{\scriptsize $\pm0.041$} & 35.7{\scriptsize $\pm{0.114}$}  \\
    Compressed & \textbf{71.6}{\scriptsize $\pm0.084$} & \textbf{90.5}{\scriptsize $\pm0.067$} & \textbf{39.7}{\scriptsize $\pm0.876$} & \textbf{75.6}{\scriptsize $\pm0.151$} & \textbf{92.7}{\scriptsize $\pm0.076$} & \textbf{34.0}{\scriptsize $\pm0.127$}  \\
    \midrule
    \multicolumn{7}{l}{\textit{ResNet-50 2x, 1000 epochs}} \\
    Uncompressed & 74.5{\scriptsize $\pm0.014$} & 92.1{\scriptsize $\pm0.031$} & 35.2{\scriptsize $\pm{0.038}$} & 77.2{\scriptsize $\pm0.057$} & 93.5{\scriptsize $\pm0.036$} & 31.8{\scriptsize $\pm{0.073}$} \\
    Compressed & \textbf{75.0}{\scriptsize $\pm0.082$} & \textbf{92.4}{\scriptsize $\pm0.086$} & \textbf{34.7}{\scriptsize $\pm0.129$} & \textbf{78.8}{\scriptsize $\pm0.088$} & \textbf{94.5}{\scriptsize $\pm0.016$} & \textbf{29.8}{\scriptsize $\pm0.028$} \\
    \bottomrule
  \end{tabular}
  \label{tab:linear_eval_r50}
\end{table}

%% file: tables/imagenet_semi_supervised.tex
\begin{table}
\small
\parbox{.5\linewidth}{
  \caption{ImageNet accuracy when training linear classifiers with 1\% and 10\% of the labeled ImageNet data, on top of frozen, self-supervised representations learned on the entire ImageNet dataset without labels.
  For the supervised baseline, the whole ResNet-50 network is trained from random initialisation. We report mean results of 3 trials.
  }
  \label{tab:semi}
  \centering
  \begin{tabular}{lcccccc}
    \toprule
    & \multicolumn{2}{c}{Top-1} & \multicolumn{2}{c}{Top-5} \\
    Method & 1\% & 10\% & 1\% & 10\% \\
    \midrule
    Supervised \cite{zhai2019s4l} & 25.4 & 56.4 & 48.4 & 80.4 \\
    \midrule
    SimCLR  & 49.3 & 63.3 & 75.8 & 85.9  \\
    C-SimCLR & \textbf{51.1} & \textbf{64.5} & \textbf{77.2} & \textbf{86.5} \\
    \midrule
    BYOL  & 55.5 & 68.2 & 79.7 & 88.4  \\
    C-BYOL & \textbf{60.6} & \textbf{70.5} & \textbf{83.4} & \textbf{90.0} \\
    \bottomrule
  \end{tabular}
}
\hfill
\parbox{.45\linewidth}{
  \caption{Comparison to other methods on ImageNet linear evaluation and supervised baselines. *: trained for 800 epochs. Other methods are 1000 epochs.}
  \label{tab:linear_eval_sota}
  \centering
  \begin{tabular}{lcc}
  \toprule
   & \multicolumn{2}{c}{ResNet-50} \\
  Method & 1x & 2x \\
  \midrule
  SimCLR \cite{chen2020simple}  & 69.3 & 74.2  \\
  SwAV* (2 crop) \cite{caron2020unsupervised,chen2020exploring}  & 71.8  & - \\
  InfoMin Aug* \cite{tian2020makes} & 73.0 & - \\
  Barlow Twins~\cite{zbontar2021barlow}  & 73.2 & - \\
  BYOL \cite{grill2020bootstrap}  & 74.3 & 77.4 \\
  \midrule 
  C-SimCLR (ours)      & 71.6 & 75.0 \\
  C-BYOL (ours)        & \textbf{75.6} & \textbf{78.8} \\
  \midrule
  Supervised \cite{chen2020simple,grill2020bootstrap} & 76.5 & 77.8 \\
  \bottomrule
  \end{tabular}
}
\end{table}

%% file: tables/robustness.tex
\begin{table}[!t]
  \scriptsize
  \caption{
  Evaluation of robustness and generalization of self-supervised models, using a ResNet-50 backbone trained on ImageNet for 1000 epochs.
  We report the mean Top-1 accuracy over 3 trials on a range of benchmarks (detailed in Sec.~\ref{sec:experiments_robustness} and the appendix) which measure an ImageNet-trained model's generalization to different domains and distributions.
  }
  \label{tab:robustness}
  \centering
  \begin{tabular}{lccccccc}
    \toprule
    Method & ImageNet-A & ImageNet-C & ImageNet-R & ImageNet-v2 & ImageNet-Vid & YouTube-BB & ObjectNet \\
    \midrule
    SimCLR	            & 1.3 &	 35.0 &	18.3 & 57.7 & 63.8 & 57.3 & 18.7 \\ 
    C-SimCLR   & \textbf{1.4} &  \textbf{36.8} & \textbf{19.6} & \textbf{58.7} & \textbf{64.7} & \textbf{59.5} & \textbf{20.8} \\ 
    \midrule
    BYOL                &   1.6 &  42.7	 & 24.4	 & 62.1 & 67.9	& 60.7	& 23.4 \\ 
    C-BYOL     &   \textbf{2.3} &  \textbf{45.1}  & \textbf{25.8}  & \textbf{63.9} & \textbf{70.8}  & \textbf{63.6}  & \textbf{25.5} \\
    \bottomrule
    \end{tabular}
    \vspace{-1.5\baselineskip}
\end{table}

%% file: tables/ceb_beta_ablation.tex
\begin{table}
  \scriptsize
  \caption{
  Ablation study of CEB compression using C-SimCLR models trained for 300 epochs with a ResNet-50 backbone.
  $\beta$ controls the level of CEB compression.
  We evaluate linear-evaluation on ImageNet, and model robustness on the remaining datasets as described in Sec.~\ref{sec:experiments_robustness}.
  }
  \centering
  \scalebox{0.95}{
      \begin{tabular}{lcccccccc}
        \toprule
        Method & ImageNet & ImageNet-A & ImageNet-C & ImageNet-R & ImageNet-v2 & ImageNet-Vid & YouTube-BB & ObjectNet \\
        \midrule
        SimCLR	      & 69.0  & \textbf{1.2} &  32.9 & 17.8 & 56.0 & 61.1 & 58.3 & 17.6 \\ 
        $\beta = 0 $      & 69.7  & 1.1 &  35.8 & 17.6 & 56.8 & 62.5 & 58.4 & 18.5 \\ 
        $\beta = 0.01 $   & 69.7  & \textbf{1.2} &  36.2 & 17.5 & 57.2 & 61.2 & 58.5 & 18.7  \\ 
        $\beta = 0.1 $    & 70.1  & 1.1 &  36.1 & 17.6 & 56.9 & 62.4 & 58.6 & 18.4 \\ 
        $\beta = 1 $      & \textbf{70.2}  & 1.1 &  \textbf{36.7} & 18.2 & \textbf{57.5} & \textbf{62.6} & \textbf{60.4} & \textbf{19.2} \\ 
        $\beta = 1.5$ & 69.7 &	1.1 & 36.4 & \textbf{18.3} &	57.3 & 62.0  &	57.9 &	18.5 \\

        \bottomrule
        \end{tabular}
    }
    \vspace{-1\baselineskip}
    \label{tab:ceb_ablation}
\end{table}

%% file: text/related_work.tex
\section{Related Work}
\label{sec:related_work}

Most methods for learning visual representations without additional annotation can be roughly grouped into three families: generative, discriminative, and bootstrapping.
Generative approaches build a latent embedding that models the data distribution, but suffer from the expensive image generation step \cite{vincent2008extracting,rezende2014stochastic,goodfellow2014generative,hinton2006fast,kingma2013auto}.
While many early discriminative approaches used heuristic pretext tasks~\cite{doersch2015unsupervised, noroozi2016unsupervised}, multi-view contrastive methods are among the recent state-of-the-art \cite{chen2020simple,he2020momentum,chen2020improved,chen2020big,oord2018representation,henaff2020data,li2021prototypical,tian2019contrastive,caron2020unsupervised}.

Some previous contributions in the multi-view contrastive family \cite{tian2020makes, zbontar2021barlow,sridharan2008information,dubois2021lossy} can be connected to the information bottleneck principle \cite{tishby2000information,tishby2015deep,alemi2016deep} but in a form of unconditional compression as they are agnostic of the prediction target, i.e. the target view in multiview contrastive learning.
As discussed in~\cite{fischer2020conditional, fischer2020ceb}, CEB performs conditional compression that directly optimizes for the information relevant to the task, and is shown theoretically and empirically better \cite{fischer2020conditional,fischer2020ceb}.
A multi-view self-supervised formulation of CEB, which C-SimCLR can be linked to, was described in \cite{fischer2020conditional}. Federici~\etal\cite{federici2020learning} later proposed a practical implementation of that,  leveraging either label information or data augmentations.
In comparison to \cite{federici2020learning}, we apply our methods with large ResNet models to well-studied large-scale classification datasets like ImageNet and study improvements in robustness and generalization, rather than using two layer MLPs on smaller scale tasks. 
This shows that compression can still work using state-of-the-art models on challenging tasks.
Furthermore, we use the vMF distribution rather than Gaussians in high-dimensional spaces, and extend beyond contrastive learning with C-BYOL.

Among the bootstrapping approaches \cite{guo2020bootstrap,caron2018deep,grill2020bootstrap} which BYOL \cite{grill2020bootstrap} belongs to, BYORL
\cite{gowal2021selfsupervised} modified BYOL \cite{grill2020bootstrap} to leverage Projected Gradient Descent \cite{madry2017towards} to learn a more adversarially robust encoder.
The focus is, however, different from ours as we concentrate on improving the generalization gap and robustness to domain shifts.

A variety of theoretical work has established that compressed representations yield improved generalization, including \cite{shamir2008learning,vera2018role,dubois2020learning}.
Our work demonstrates that these results are valid in practice, for important problems like ImageNet, even in the setting of self-supervised learning.
Our theoretical analysis linking Lipschitz continuity to compression also gives a different way of viewing the relationship between compression and generalization, since smoother models have been found to generalize better (e.g., \cite{bruna2013invariant}).
Smoothness is particularly important in the adversarial robustness setting~\cite{weng2018evaluating,fazlyab2019efficient,yang2020closer}, although we do not study that setting in this work.

%% file: text/conclusion.tex
\section{Conclusion}
\label{sec:conclusion}

We introduced compressed versions of two state-of-the-art self-supervised algorithms, SimCLR~\cite{chen2020simple} and BYOL~\cite{grill2020bootstrap}, using the Conditional Entropy Bottleneck (CEB)~\cite{fischer2020ceb}. 
Our extensive experiments verified our hypothesis that compressing the information content of self-supervised representations yields consistent improvements in both accuracy and robustness to domain shifts.
These findings were consistent for both SimCLR and BYOL across different network backbones, datasets and training schedules.
Furthermore, we presented an alternative theoretical explanation of why C-SimCLR models are more robust, in addition to the information-theoretic view~\cite{fischer2020conditional, fischer2020ceb, achille2017emergence, achille2018information}, by connecting Lipschitz continuity to compression.

\paragraph{Limitations.}
We note that using CEB often requires explicit and restricted distributions.
This adds certain constraints on modeling choices.
It also requires additional effort to identify or create required random variables, and find appropriate distributions for them.
Although we did not need additional trainable parameters for C-SimCLR, we did for C-BYOL, where we added a linear layer to the online encoder, and a 2-layer MLP to create $b(\cdot|y)$.
It was, however, not difficult to observe the von Mises-Fisher distribution corresponds to loss function of BYOL and SimCLR, as well as other recent InfoNCE-based contrastive methods \cite{caron2020unsupervised,chen2020improved,he2020momentum}.

\paragraph{Potential Negative Societal Impact.}
Our work presents self-supervised methods for learning effective and robust visual representations.
These representations enable learning visual classifiers with limited data (as shown by our experiments on ImageNet with 1\% or 10\% training data), and thus facilitates applications in many domains where annotations are expensive or difficult to collect.

Image classification systems are a generic technology with a wide range of potential applications.
We are unaware of all potential applications, but are cognizant that each application has its own merits and societal impacts depending on the intentions of the individuals building and using the system.
We also note that training datasets contain biases that may render models trained on them unsuitable for certain applications.
It is possible that people use classification models (intentionally or not) to make decisions that impact different groups in society differently.

%% file: text/appendix.tex
We provide additional implementation details in \Cref{sec:impl_detail}, details of our linear evaluation protocol in \Cref{sec:detail_eval}, experiments on transfer of the compressive representations to other classification tasks in \Cref{sec:transfer_classification_tasks}, and further ablations in \Cref{sec:further_ablations}.
We provide further details about our robustness evaluation in \Cref{sec:robustness_benchmark_details}.
Finally, we provide a more detailed explanation of the relation between Lipschitz continuity and SimCLR with CEB compression, introduced in \Cref{sec:lipschitz} of the main paper, in \Cref{sec:complete_lipschitz}.

\section{Implementation details and hyperparameters}
\label{sec:impl_detail}
In this section, we further describe our implementation details.
Our implementation is based off of the public implementation of SimCLR~\cite{chen2020simple}.
In general, we closely follow the design choices of BYOL~\cite{grill2020bootstrap} for both of our SimCLR and BYOL implementations.
Despite having different objectives, BYOL and SimCLR share many components, including image augmentations, network architectures, and optimization settings.
As explained in the original paper \cite{grill2020bootstrap}, BYOL itself may be considered as a modification to SimCLR with a slow moving average target network, an additional predictor network, and switching the learning target from InfoNCE to a regression loss.
Therefore, many of the design choices and hyperparameters are applicable to both.
As explained in \Cref{sec:experiments_setup}, we align SimCLR with BYOL on the choices of image augmentations, network architecture, and optimization settings in order to reduce the number of variables in comparison.

\subsection{Image augmentations}
\label{sec:detail_aug}
During self-supervised training, we use the set of image augmentations from BYOL~\cite{grill2020bootstrap} for all our models.

\begin{itemize}
    \item Random cropping: randomly select a patch of the image, with an area uniformly sampled between 8\% and 100\% of that original image, and an aspect ratio logarithmically sampled between $3/4$ and $4/3$.
    Then this patch is resized to $224 \times 224$ using bicubic interpolation.
    \item Left-to-right flipping: randomly flip the patch.
    \item Color jittering: brightness, contrast, saturation and hue of an image are shifted by a uniformly random offset.
    The order to apply these adjustments is randomly selected for each patch. 
    \item Color dropping:
    RGB pixel values of an image are converted to grayscale according to $0.2989r + 0.5870g + 0.1140b$.
    \item Gaussian blurring: We use a $23 \times 23$ kernel to blur the $224 \times 224$ image, with a standard deviation uniformly sampled over $[0.1, 2.0]$.
    \item Solarization: This is a color transformation $x = x \cdot \textbf{1}_{\{x < 0.5\}} + (1-x) \cdot \textbf{1}_{\{x\geq0.5\}}$ for pixels with values in $[0, 1]$ (we convert all pixel values into floats between $[0, 1]$).
\end{itemize}

As described in Sec.~\ref{sec:methods}, we use augmentation functions $t$ and $t'$ to transform an image into two views. 
$t$ and $t'$ are compositions of the above image augmentations in the listed order, each applied with a predefined probability.
The image augmentation parameters to generate $t$ and $t'$ are listed in Table~\ref{tab:image_aug}.

During evaluation, we perform center cropping, as done in \cite{chen2020simple,grill2020bootstrap}.
Images are resized to 256 pixels along the shorter side, after which a $224 \times 224$ center crop is applied.
During both training and evaluation, we normalize image RGB values by subtracting the average color and dividing by the standard deviation, computed on ImageNet, after applying the augmentations.

\paragraph{Differences from the original SimCLR~\cite{chen2020simple}.}
Since the image augmentation parameters that BYOL~\cite{grill2020bootstrap} uses are different from the original SimCLR, we list the original SimCLR parameters in the last column of \Cref{tab:image_aug}, which are the same for $t$ and $t'$, to clarify the differences.
Additionally, the original SimCLR samples the aspect ratio of cropped patches uniformly, instead of logarithmically, between $[3/4, 4/3]$.

\begin{table}
  \caption{Image augmentation parameters. We use the hyperparameter values from BYOL~\cite{grill2020bootstrap}, and include the values from the original SimCLR~\cite{chen2020simple} as reference.}
  \label{tab:image_aug}
  \centering
  \begin{tabular}{lccc}
    \toprule
    Parameter & $t$ & $t'$ & Orig. SimCLR \cite{chen2020simple} \\
    \midrule
    Random crop probability & 1.0 & 1.0 & 1.0 \\  
    Flip probability & 0.5 & 0.5 & 0.5 \\
    Color jittering probability & 0.8 & 0.8 & 0.8 \\
    Brightness adjustment max strength & 0.4 & 0.4 & 0.8 \\
    Contrast adjustment max strength & 0.4 & 0.4 & 0.8 \\
    Saturation adjustment max strength & 0.2 & 0.2 & 0.8 \\
    Hue adjustment max strength & 0.1 & 0.1 & 0.2 \\
    Color dropping probability & 0.2 & 0.2 & 0.2 \\
    Gaussian blurring probability & 1.0 & 0.1 & 1.0 \\
    Solarization probability & 0.0 & 0.2 & 0.0 \\
    \bottomrule
  \end{tabular}
\end{table}

\subsection{Network architecture}
\label{sec:detail_net_arch}
Following \cite{chen2020simple,grill2020bootstrap}, we use ResNet-50 \cite{he2016deep} as our backbone convolutional encoder (the ``Conv'' part in \Cref{fig:ceb_simclr} and \Cref{fig:byol}).
We vary the ResNet width \cite{zagoruyko2016wide} (and thus the number of parameters) from 1$\times$ to 2$\times$.
In Sec.~\ref{sec:cbyol_w_deeper_resnets}, we additionally report C-BYOL results with different ResNet depth, from 50 to 152.
The representations $h_x, h_y$ in SimCLR and $h, h_t, h_t'$ in BYOL correspond to the 2048-dimensional (for ResNet-50 1$\times$) final average pooling layer output.
These representations are projected to a smaller space by an MLP (called ``projection`` in \Cref{fig:ceb_simclr} and \Cref{fig:byol}).
As in \cite{grill2020bootstrap}, this MLP consists of a linear layer with output size 4096 followed by batch normalization~\cite{ioffe2015batch}, ReLU, and a final linear layer with output dimension 256.
$q(\cdot)$ in BYOL/C-BYOL (Fig.~\ref{fig:byol}) is called the predictor.  %
The predictor $q(\cdot)$ is also a two-layer MLP which shares the same architecture with the projection MLP \cite{grill2020bootstrap}.

\paragraph{Differences from the original SimCLR~\cite{chen2020simple}.}
The original SimCLR \cite{chen2020simple} uses a 2048-d hidden layer and a 128-d output layer for the projection MLP, after which an additional batch normalization is applied to the 128-d output.
Both BYOL \cite{grill2020bootstrap} and our work do not have this batch normalization on the last layer. 
We did not observe significant change in performance for the uncompressed models and found it harmful to the compressed models.

\subsection{von Mises-Fisher Distributions}
We use the vMF implementation in public Tensorflow Probability (TFP) library \cite{dillon2017tensorflow}, specifically the current TFP version 0.13. 
We have found that sampling and computing log probabilities in high dimensions with the current TFP version has become sufficiently stable and fast to train all of the models in our paper.\footnote{Previous versions of TFP were unstable for sampling from vMF distributions with higher than 5 dimensions, and at the time of writing, the authors of the library have not updated the documentation to indicate that this is no longer the case.} 

\subsection{Optimization}
\label{sec:detail_training}
We follow BYOL \cite{grill2020bootstrap} for our optimization settings.
During self-supervised training, we use the LARS optimizer~\cite{you2017large} with a cosine decay learning rate schedule \cite{loshchilov2016sgdr} over 1000 epochs and a linear warm-up period at the beginning of training.
The linear warm-up period is 10 epochs in most cases.
We increase it to 20 epochs for BYOL and C-BYOL with larger ResNets (ResNet-50 2x, ResNet-101, ResNet-152) as we found it helpful to prevent mode collapse and improve performance.
In most cases, we set the base learning rate to 0.2 and scale it linearly by batch size (LearningRate = 0.2 $\times$ BatchSize/256).
For C-BYOL, we increase the base learning rate to 0.26 for better performance.
For careful comparison, we extensively searched base learning rate for BYOL but did not find a configuration better than 0.2 as used in the original work~\cite{grill2020bootstrap}.
We use a weight decay of $1.5\times10^{-6}$.
For the BYOL/C-BYOL target network, the exponential moving average update rate $\alpha$ starts from $\alpha_{\text{base}}=0.996$ and ramps up to 1 with a cosine schedule, $\alpha \triangleq 1-(1-\alpha_{\text{base}})(\cos(\pi k/K)+1)/2$ where $k$ is the current training step and $K$ is the total number of training steps.

For 300-epoch models used in ablations, we set the base learning rate to 0.3 in most cases, and increase it to 0.35 for C-BYOL.
We use a weight decay of $10^{-6}$.
For BYOL and C-BYOL, the base exponential moving average update rate $\alpha_{\text{base}}$ is set to 0.99.

We note that there is a small chance that both BYOL and C-BYOL can end up with collapsed solutions, but it mostly happens in early phase of training and is easy to observe with the learning objective spiking or reaching NaN.

\paragraph{Differences from the original SimCLR~\cite{chen2020simple}.}
Optimization settings of the original SimCLR are very similar but, for 1000-epoch training, they use a base learning rate of 0.3 and weight decay of $10^{-6}$.

\subsection{SimCLR and C-SimCLR details}
\label{sec:detail_simclr}
As described in \Cref{sec:detail_aug}, \Cref{sec:detail_net_arch}, \Cref{sec:detail_training}, we made minor modifications to the original SimCLR to align with BYOL on the choices of image augmentations, network architecture, and optimization settings.
With these modifications, our SimCLR baseline reproduction outperforms the original (top-1 accuracy 70.6\% versus 69.3\%).

For C-SimCLR, we use $\kappa_e=1024$ for the true encoder $e(\cdot|x)$ and $\kappa_b=10$ for the backward encoder, where $e(\cdot|x)$ and $b(\cdot|y)$ are von Mises-Fisher distributions.
The compression factor $\beta$ that we use for C-SimCLR is $1.0$.
Note that the original SimCLR has temperature $\tau=0.1$ which is equivalent to having $\kappa_b=10$, since $\kappa_b=1/\tau$.

\subsection{BYOL and C-BYOL details}
\label{sec:detail_byol}

\begin{table}
  \caption{Ablation study on BYOL models trained for 300 epochs. Top-1 denotes the linear evaluation Top-1 accuracy on ImageNet.}
  \label{tab:byol_300e_ablation}
  \centering
  \begin{tabular}{lc}
    \toprule
    Method & Top-1 \\
    \midrule
    BYOL $w_{\text{byol}}=1.0$ & 72.5 \\  
    BYOL $w_{\text{byol}}=5.0$ & 72.8 \\
    BYOL $w_{\text{byol}}=5.0$ + 256-d linear layer + $l_2$-normalization & 72.8 \\
    BYOL $w_{\text{byol}}=5.0$ + 256-d linear layer + $l_2$-normalization + sampling & 72.8 \\
    C-BYOL $w_{\text{byol}}=5.0$ &  73.6 \\
    \bottomrule
  \end{tabular}
  
  \caption{The effect of loss weights on BYOL models trained for 1000 epochs. Top-1 denotes the linear evaluation Top-1 accuracy on ImageNet.}
  \label{tab:byol_1000e_loss_w}
  \centering
  \begin{tabular}{lc}
    \toprule
    Method & Top-1 \\
    \midrule
    BYOL $w_{\text{byol}}=1.0$ & 74.2  \\  
    BYOL $w_{\text{byol}}=2.0$ & 74.2 \\
    BYOL $w_{\text{byol}}=5.0$ & 74.2 \\
    \bottomrule
  \end{tabular}
\end{table}

As shown in \Cref{tab:byol_300e_ablation} and \Cref{tab:byol_1000e_loss_w}, our BYOL implementation stably reproduces results comparable to \cite{grill2020bootstrap} with 300 and 1000 epochs of training.
An interesting behavior we observed is that, for shorter training with 300 epochs, scaling the BYOL regression loss can improve performance.
Specifically we add a weight multiplier $w_{\text{byol}} = \kappa_d/2$ to the BYOL loss Eq.~\eqref{eq:byol_loss}.
\begin{align}
    L_{\text{byol}} = w_{\text{byol}}||\mu_e - y'||^2_2
\label{eq:scaled_byol_loss}
\end{align}
\Cref{tab:byol_300e_ablation} shows that multiplying the loss by five increases the linear evaluation accuracy from 72.5\% to 72.8\%. 
This improvement is consistent across multiple runs.
Therefore, we choose $w_{\text{byol}}=5$ for 300-epoch BYOL/C-BYOL.
However, we do not see the same improvement for 1000-epoch models.
\Cref{tab:byol_1000e_loss_w} shows that $w_{\text{byol}}$ makes little difference for 1000-epoch BYOL models.
We still choose $w_{\text{byol}}=2$ for all 1000-epoch BYOL and C-BYOL models since it tends to work better than $w_{\text{byol}}=1$ for the compressed models and models with larger ResNets.

Furthermore, \Cref{tab:byol_300e_ablation} verifies that the additional linear layer with $l_2$-normalization that we added for C-BYOL and $z$ sampling (both were described in \Cref{sec:byol}) do not result in a difference in performance.
The improvement happens only when CEB compression is used.

We set $\kappa_e=16384.0$, $\kappa_b=10.0$, and the compression factor $beta=1.0$ for C-BYOL if not specified otherwise.

\section{Linear evaluation protocol on ImageNet}
\label{sec:detail_eval}

As common in self-supervised learning literature~\cite{grill2020bootstrap,chen2020simple,kolesnikov2019revisiting,chen2020improved}, we assess the performance of our representations learned on the ImageNet training set (without labels) by training a linear classifier on top of the frozen representations using the labeled data.
For training this linear classifier, we only apply the random cropping and flipping image augmentations.
We optimize the cross-entropy loss using SGD with Nesterov momentum over 40 epochs.
We use a batch size of 1024 and momentum of 0.9 without weight decay, and sweep the base learning rate over $\{0.4, 0.3, 0.2, 0.1, 0.05\}$ to choose the best learning rate on a validation set, following \cite{grill2020bootstrap}.
We perform center cropping during evaluation, as done in \cite{chen2020simple,grill2020bootstrap}.
Images are resized to 256 pixels along the shorter side, after which the $224 \times 224$ center crop is selected.
During both training and evaluation, we normalize image RGB values by subtracting the average color and dividing by the standard deviation, computed on ImageNet, after applying the augmentations.

\paragraph{Learning with a few labels}
In \Cref{sec:experiments_imagenet} we described learning the linear classifier on 1\% and 10\% of the ImageNet training set with labels.
We performed this experiment with the same 1\% and 10\% splits from \cite{chen2020simple}.

\section{Transfer to other classification tasks}
\label{sec:transfer_classification_tasks}

\begin{table}
  \scriptsize
  \caption{
    Transfer to other classification tasks, by performing linear evaluation.
    The backbone network is ResNet-50, pretrained in a self-supervised fashion for 1000 epochs.
  }
  \label{tab:classification_transfer}
  \centering
  \scalebox{0.925}{
  \begin{tabular}{lcccccccccccc}
        \toprule
        Method & Food101 & CIFAR10 & CIFAR100 & Flowers & Pet & Cars & Caltech-101 & DTD & SUN397 & Aircraft & Birdsnap \\
        \midrule
        SimCLR    & 72.5 & 91.1 & 74.4 & 88.4 & 83.5 & 49.7 & 89.5 & 72.5 & 61.8 & 51.6 & 35.4 \\
        C-SimCLR  & \textbf{73.0} & \textbf{91.6} & \textbf{75.2} & \textbf{89.0} & \textbf{84.0} & \textbf{52.7} & \textbf{91.2} & \textbf{73.0} & \textbf{62.3} & \textbf{53.5} & \textbf{38.2} \\
        \bottomrule
      \end{tabular}
     }
\end{table}

We analyze the effect of compression on transfer to other classification tasks in Table~\ref{tab:classification_transfer}.
This allows us to assess whether the compressive representations learned by our method are generic and transfer across image domains.

\paragraph{Datasets.}
We perform the transfer experiments on the Food-101 dataset~\cite{bossard2014food}, CIFAR-10 and CIFAR-100 \cite{krizhevsky2009learning}, Birdsnap~\cite{berg2014birdsnap}, SUN397~\cite{xiao2010sun}, Stanford Cars~\cite{krause2013collecting}, FGVC Aircraft~\cite{maji2013fine}, the Describable Texture Dataset (DTD) \cite{cimpoi2014describing}, Oxford-IIIT Pets~\cite{parkhi2012cats}, Caltech-101~\cite{fei2004learning}, and Oxford 102 Flowers~\cite{nilsback2008automated}.
We carefully follow their evaluation protocol, i.e. we report top-1 accuracy for Food-101, CIFAR-10, CIFAR-100, Birdsnap, SUN397, Stanford Cars, anad DTD; mean per-class accuracy for FGVC Aircraft, Oxford-IIIT Pets, Caltech-101, and Oxford 102 Flowers.
These datasets are also used by~\cite{chen2020simple, grill2020bootstrap, kornblith2019better}.
More exhaustive details about train, validation, and test splits of these datasets can be found in Section D of~\cite{grill2020bootstrap} (arXiv v3).

\paragraph{Transfer via linear classifier.}
To demonstrate the effectiveness of compressed representations, we compare SimCLR and C-SimCLR representations as an example.
We freeze the representations of our model and train an $\ell_2$-regularized multinomial logitstic regression classifier on top of these frozen representations.
We minimize the cross-entropy objective using the L-BFGS optimizer.
As in \cite{grill2020bootstrap,chen2020simple}, we selected the $\ell_2$ regularization parameter from a range of 45 logarithmically spaced values between $[10^{-6}, 10^5]$.

We observe in Table~\ref{tab:classification_transfer} that our Compressed SimCLR model consistently outperforms the uncompressed SimCLR baseline on each of the 11 datasets we tested.
We note absolute improvements in accuracy ranging from 0.5\% (CIFAR-10, SUN397) to 3\% (Stanford Cars).
These experiments suggest that the representations learned by compressed model are generic, and transfer beyond the ImageNet domain which they were learned on.

\section{Extra C-BYOL results with Deeper ResNets}
\label{sec:cbyol_w_deeper_resnets}

\begin{table}[t]
  \caption{C-BYOL and BYOL trained for 1000 epochs with different ResNet depth. We report ImageNet Top-1 accuracy from linear evaluation, averaged over 3 trials.}
  \label{tab:cbyol_w_deeper_resnets}
  \centering
  \begin{tabular}{lcccc}
    \toprule
    & \multicolumn{2}{c}{C-BYOL} & \multicolumn{2}{c}{BYOL \cite{grill2020bootstrap}} \\
    Architecture & Top-1 & Top-5 & Top-1 & Top-5 \\
    \midrule
    ResNet-50 & \textbf{75.6} & \textbf{92.7} & 74.3 & 91.6  \\
    ResNet-101 & \textbf{77.8} & \textbf{93.9} & 76.4 & 93.0 \\
    ResNet-152 & \textbf{78.7} & \textbf{94.4} & 77.3 & 93.7 \\
    \bottomrule
  \end{tabular}
\end{table}

In \Cref{tab:cbyol_w_deeper_resnets}, we additionally report results of C-BYOL and BYOL retrained for 1000 epochs with ResNet-101 and ResNet-152, as it could be of interest to demonstrate improvements over the state-of-the-art BYOL on these deeper ResNet models. 
It can be observed that C-BYOL gives significant gains across ResNets with different depths.

\section{Additional Ablations}
\label{sec:further_ablations}

The hyperparameter and architecture choices of SimCLR and BYOL have been investigated in the original works \cite{chen2020simple,grill2020bootstrap}. 
Here we focus on analysing hyperparameters specific to C-SimCLR and C-BYOL.

\Cref{tab:kappa_e_csimclr}, \Cref{tab:kappa_e_cbyol} and \Cref{tab:kappa_b} show how changing $\kappa_e$ and $\kappa_b$ affect the results, respectively.
We also investigate the effect of the compression factor $\beta$ for C-BYOL (\Cref{tab:cbyol_beta}) in addition to the compression analysis for SimCLR in Sec.~\ref{sec:experiments_ablations}.
Similar to C-SimCLR, as compression strength ($\beta$) increases, the linear evaluation result improves, with $\beta=1.0$ obtaining the best results, but overly strong compression leads to a drop in performance.

Finally, we conduct a preliminary exploration on the interplay between CEB compression and image augmentations, using cropping area ratio as an example in \Cref{tab:area_range_lower_bound}.
As described in \Cref{sec:detail_aug}, we follow \cite{grill2020bootstrap,chen2020simple} to randomly crop an image to an area between 8\% and 100\% of the original image.
We refer to this 8\% as the ``area lower bound``, which is the most aggressive cropping area ratio that can happen.
As the area lower bound decreases, we are reducing the amount of information that can be shared between the two representations, because there is less and less mutual information between the two images: $I(X;X')$ gets smaller the more we reduce the area lower bound \cite{tian2020makes}.
Thus, smaller area lower bounds should force the model to be more compressed.
What we see in \Cref{tab:area_range_lower_bound} is that the SimCLR models are much more sensitive to the changes in the area lower bound than the C-SimCLR models are.
We speculate that this is because the compression done by the C-SimCLR objective overlaps to some extent with the compression given by varying the area lower bound.
Regardless, the compression due to the area lower bound hyperparameter appears to be insufficient to adequately compress away irrelevant information in the SimCLR model, which is why the C-SimCLR models continue to outperform the SimCLR models at all area lower bound values.

\begin{table}[t]
  \caption{The effect of varying $\kappa_e$ for C-SimCLR models. We report ImageNet Top-1 accuracy from linear evaluation.}
  \label{tab:kappa_e_csimclr}
  \centering
  \begin{tabular}{rcccccc}
    \toprule
    $\kappa_e$ & 256 & 512 & 1024 & 2048 & 4096 & 8192 \\
    \midrule
    ImageNet Top-1 accuracy & 69.8 & 69.8 & 70.2 & 69.8 & 69.6 & 69.6 \\
    \bottomrule
  \end{tabular}
  \caption{The effect of varying $\kappa_e$ for C-BYOL models. We report ImageNet Top-1 accuracy from linear evaluation.}
  \label{tab:kappa_e_cbyol}
  \centering
  \begin{tabular}{rcccc}
    \toprule
    $\kappa_e$ & 4096 & 8192 & 16384 & 32768 \\
    \midrule
    ImageNet Top-1 accuracy & 73.0 & 73.3 & 73.6 & 73.2 \\
    \bottomrule
  \end{tabular}

  \caption{The effect of varying $\kappa_b$ for C-SimCLR and C-BYOL models. We report ImageNet Top-1 accuracy from linear evaluation.}
  \label{tab:kappa_b}
  \centering
  \begin{tabular}{lccccc}
    \toprule
    Method & $\kappa_b=$1 & 3 & 10 & 15 & 20 \\
    \midrule
    C-SimCLR & 65.0 & 68.5 & 70.2 & 69.1 & 68.6 \\
    C-BYOL & 73.1 & 73.3 & 73.6 & 73.4 & 73.2 \\
    \bottomrule
  \end{tabular}

  \caption{The effect of $\beta$ on C-BYOL. Note that \Cref{tab:ceb_ablation} in the main paper studied this effect on C-SimCLR.
  The final column is the uncompressed BYOL baseline.
  }
  \label{tab:cbyol_beta}
  \centering
  \begin{tabular}{rccccc}
    \toprule
    $\beta$ & 1.5 & 1.0 & 0.1 & 0.01 & BYOL \\
    \midrule
    ImageNet Top-1 accuracy & 73.4 & 73.6 & 73.1 & 73.0 & 72.8 \\
    \bottomrule
  \end{tabular}

  \caption{The effect of varying the area range lower rounds for SimCLR and Compressed SimCLR. 
  We report the ImageNet Top-1 accuracy from linear evaluation.
  Note how the baseline SimCLR model is much more sensitive to this data-augmentation hyperparameter.
  }
  \label{tab:area_range_lower_bound}
  \centering
  \begin{tabular}{lcccc}
    \toprule
    Method & 8\% & 16\% & 25\% & 50\%\\
    \midrule
    SimCLR & 69.0 & 68.6 & 67.6 & 61.4  \\
    C-SimCLR & 70.2 & 70.0 & 68.9 & 64.3 \\
    \bottomrule
  \end{tabular}
\end{table}

\section{Robustness benchmark details}
\label{sec:robustness_benchmark_details}
In this section, we provide some additional details on each of the datasets used in our robustness evaluations.
Note that we use the public robustness benchmark evaluation code of~\cite{djolonga2020robustness, djolonga2020robustnesscode}.\footnote{\url{https://github.com/google-research/robustness_metrics}}

ImageNet-A~\cite{hendrycks2021natural}: 
This dataset of ``Natural adversarial examples'' consists of images of ImageNet classes which a ResNet-50 classifier failed on.
The dataset authors performed manual, human-verification to ensure that the predictions of the ResNet-50 model were indeed incorrect and egregious~\cite{hendrycks2021natural}.

ImageNet-C~\cite{hendrycks2019benchmarking}: 
This dataset adds 15 corruptions to ImageNet images, each at 5 levels of severity.
We report the average accuracy over all the corruptions and severity levels.

ImageNet-R~\cite{hendrycks2020many}: 
This dataset, which has the full name ``Imagenet Rendition'',  captures naturally occuring distribution changes in image style, camera operation and geographic location.

ImageNet-v2~\cite{recht2019imagenet}: 
This is a new test set for ImageNet, and was collected following the same protocol as the original ImageNet dataset.
The authors posit that the collected images are more ``difficult'', and observed consistent accuracy drops across a wide range of models trained on the original ImageNet.

ObjectNet~\cite{barbu2019objectnet}: This is a more challenging test set for ImageNet, where the authors control for different viewpoints, backgrounds and rotations.
Note that ObjectNet has a vocabulary 313 object classes, of which 113 are common with ImageNet.
Following~\cite{djolonga2020robustnesscode}, we evaluate on only the images in the dataset which have one of the 113 ImageNet labels.
Our network is still able to predict any one of the 1000 ImageNet classes though.

ImageNet-Vid and YouTube-BB~\cite{shankar2019image} evaluate the robustness of image classifiers to natural perturbations arising in video.
This dataset was created by~\cite{shankar2019image} by augmenting the ImageNet-Vid~\cite{russakovsky2015imagenet} and YouTube-BB~\cite{real2017youtubeboundingboxes} datasets with additional annotations.

\section{Analysis of Lipschitz Continuity and Compression}
\label{sec:complete_lipschitz}

In this section, we provide a more detailed explanation of the relation between Lipschitz continuity and SimCLR with CEB compression, introduced in \Cref{sec:lipschitz}.
Lipschitz Continuity provides a way of measuring how smooth a function is.
For some function $f$ and a distance measure $D(f(x_1), f(x_2))$, Lipschitz continuity defines an upper bound on how quickly $f$ can change as $x$ changes:
\begin{align}
L ||\Delta x|| \geq D(f(x), f(x + \Delta x))
\end{align}
where $L$ is the Lipschitz constant, $\Delta x$ is the vector change in $x$, and $||\Delta x|| > 0$.

Frequently, the choice of $D$ is the absolute difference function: $|f(x_1) - f(x_2)|$.
However, we can use a multiplicative distance rather than an additive distance by considering the absolute difference of the logs of the functions:
\begin{align}
\label{eq:multiplicative_distance}
D(f(x_1), f(x_2)) \equiv | \log f(x_1) - \log f(x_2) |
\end{align}

It is trivial to see that \Cref{eq:multiplicative_distance} obeys the triangle inequality, which can be written:
\begin{align}
\label{eq:triangle_ineq}
|a - b| \geq |a| - |b|
\end{align}
\Cref{eq:triangle_ineq} is true for any scalars $a$ and $b$.
Setting $a = \log f(x_1)$ and $b = \log f(x_2)$ is sufficient, given that $f(\cdot)$ is a positive, scalar-valued function.
For $D(\cdot)$ to be a valid distance metric, $f(x)$ must also satisfy the identity of indiscernibles requirement: $f(x_1) = f(x_2) \Leftrightarrow x_1 = x_2$.
If that requirement is violated, then $D(\cdot)$ becomes a pseudometric, which is inconsistent with Lipschitz continuity.

Noting that $|a - b| \equiv \max(a - b, b - a)$, we will simplify the analysis by considering the two arguments to the implicit $\max$ in \Cref{eq:multiplicative_distance} one at a time, starting with:
\begin{align}
L &\geq  \frac{1}{||\Delta x||} ( \log f(x) - \log f(x + \Delta x) )
\end{align}

If we define $f(x)$ to be our encoder distribution, $e(z|x)$, we get a function of $z$ of Lipschitz value:\footnote{%
  Note that if we choose an encoder distribution where the density ever goes to 0 or $\infty$, \Cref{eq:encoder_lipschitz} will have a maximum value of $\infty$.
  Of course, it's generally easy to avoid this situation by choosing ``well-behaved'' distributions like the Gaussian or von Mises-Fisher distributions, whose densities are non-zero on the entire domain, and to parameterize them with variance or concentration parameters that don't go to 0 or $\infty$, respectively.
}
\begin{align}
\label{eq:encoder_lipschitz}
L(z) \geq \frac{1}{||\Delta x||} ( \log e(z|x) - \log e(z|x + \Delta x) )
\end{align}
Note that the encoder distribution must not violate the identity of indiscernibles property: $\forall z: e(z|x_1) = e(z|x_2) \Leftrightarrow x_1 = x_2$.
This is not the case in general, but for reasonable distribution families, the sets of $z$ that violate this property for any $(x_1, x_2)$ pair will have measure zero.
In the case that $e(z|\cdot)$ is parameterized by some function $f(\cdot)$, such as a neural network, $f$ must also not violate the identity of indiscernibles property.
This argues in favor of using invertible networks for $f$, or at least not using activation functions like relu that are likely to cause $f$ to map multiple $x$ values to some constant.
We note that in practice, it doesn't seem to matter, as shown empirically in \Cref{sec:experiments}.

As $e(z|x)$ is parameterized by the output of our model, \Cref{eq:encoder_lipschitz} captures the semantically relevant smoothness of the model.
For example, if our encoder distribution is a Gaussian with learned mean and variance, the impact of the model parameters on the means is semantically distinct from the impact of the model parameters on the variance.
In that setting, using the parameter vectors themselves naively in a Lipschitz formulation like $L_{\text{naive}} ||\Delta x|| \geq || f_\theta(x) - f_\theta(x + \Delta x) ||$, where $f_\theta$ outputs concatenated mean and variance parameters, would clearly fail to correctly capture the model's smoothness.
Our formulation using the encoder \emph{distribution} directly does not have this flaw, and thus generalizes to capture a notion of smoothness for any choice of distribution parameterization.
Note that this notion of smoothness of the distribution still depends directly on the the smoothness of the underlying function that generates the distribution's parameters, while also capturing the smoothness of the distribution itself.

We can remove the dependence on $z$ of \Cref{eq:encoder_lipschitz} by taking the expectation over $z$ with respect to $e(z|x)$.
This gives us an \emph{expected} Lipschitz value:
\begin{align}
\mathbb{E}_{z \sim e(z|x)} L(z) &\geq \frac{1}{||\Delta x||} \mathbb{E}_{z \sim e(z|x)} \log e(z|x) - \log e(z|x + \Delta x) \\
&= \frac{1}{||\Delta x||} \operatorname{KL}[ e(z|x) || e(z|x + \Delta x) ] \label{eq:lipschitz_kl}
\end{align}
It is important to note that \Cref{eq:lipschitz_kl} no longer obeys the triangle inequality, due to the $\operatorname{KL}$ divergence, since it is easy to find three distributions $p, q, r$ such that $\operatorname{KL}[p||q] > \operatorname{KL}[p||r] + \operatorname{KL}[q||r]$.

We could also have computed the expectation over $z \sim e(z|x + \Delta x)$, yielding:
\begin{align}
\mathbb{E}_{z \sim e(z|x + \Delta x)} L(z) &\geq -\frac{1}{||\Delta x||} \operatorname{KL}[ e(z|x + \Delta x) || e(z|x) ]
\end{align}
But this is trivially true due to $L(z)$ being non-negative and the negative $\operatorname{KL}$ term being non-positive, so we can ignore this term here.
However, when we consider the second argument to the implicit $\max$ in \Cref{eq:multiplicative_distance}, the negative and positive $\operatorname{KL}$ terms are swapped, and we are left with:
\begin{align}
\mathbb{E}_{z \sim e(z|x + \Delta x)} L(z) &\geq \frac{1}{||\Delta x||} \operatorname{KL}[ e(z|x + \Delta x) || e(z|x) ]
\end{align}

When we take the expectations over $z$, the resulting $\operatorname{KL}$ divergences have an underlying quadratic growth in $||\Delta x||$: as $||\Delta x||$ increases linearly, the $\operatorname{KL}$ divergences increase quadratically.\footnote{%
  This is easiest to see with Gaussian distributions whose means are parameterized by an identity map of $x$ and $x + \Delta x$: the $\operatorname{KL}$ divergence is quadratic in difference of the means, which is $||\Delta x||$.
}
This is why the $\operatorname{KL}$ divergence violates the triangle inequality, and also why it is problematic for measuring Lipschitz continuity: in general, $L$ will be unbounded when measured by the $\operatorname{KL}$ even when the underlying function $f(x)$ parameterizing the distributions has a bounded Lipschitz constant, since the $\operatorname{KL}$ will always grow faster then $|| \Delta x ||$.
We can address this by instead considering the squared Lipschitz constant:
\begin{align}
L^2 ||\Delta x||^2 \geq \operatorname{KL}[ e(z|x) || e(z|x + \Delta x) ]
\quad \text{and} \quad L^2 ||\Delta x||^2 \geq \operatorname{KL}[ e(z|x + \Delta x) || e(z|x) ]
\end{align}
which is equivalent to:
\begin{align}
L^2 \geq \frac{1}{||\Delta x||} \mathbb{E}_{z \sim e(z|x)} L(z)
\quad \text{and} \quad L^2 \geq \frac{1}{||\Delta x||} \mathbb{E}_{z \sim e(z|x + \Delta x)} L(z)
\end{align}

Finally, we note the following relationship:
\begin{align}
L^2 = \max_{x,\Delta x} \max\left( \frac{1}{||\Delta x||^2} \operatorname{KL}[ e(z|x) || e(z|x + \Delta x) ],  \frac{1}{||\Delta x||^2} \operatorname{KL}[ e(z|x + \Delta x) || e(z|x) ] \right)
\end{align}
In words, the true squared Lipschitz constant of the encoder is equal to the least smooth $(x,\Delta x)$ pair, as measured by the greater of the two $\operatorname{KL}$ divergences at that pair.

Putting all of this together, we observe that the following two $\operatorname{KL}$ divergences together give a lower bound on the encoder's Lipschitz constant:
\begin{align}
\label{eq:lipschitz_log_bound}
L^2 \geq \frac{1}{||\Delta x||^2} \max \Big( \operatorname{KL}[ e(z|x) || e(z|x + \Delta x) ],\, \operatorname{KL}[ e(z|x + \Delta x) || e(z|x) ] \Big)
\end{align}
Thus, taking the pointwise maximum across pairs of inputs in any dataset gives a valid estimate of the maximum lower bound of the encoder's Lipschitz constant.
\Cref{eq:lipschitz_log_bound} can be evaluated directly on any pair of valid inputs $(x, x + \Delta x)$.
\Cref{eq:lipschitz_log_bound} is the same as \Cref{eq:lipschitz_log_bound1} used in \Cref{sec:lipschitz}.

\paragraph{Example: the von Mises-Fisher distribution.}
An exponential family distribution has the form:
\begin{align}
    h(z) \exp(\eta^T T(z)-A(\eta))
\end{align}
where $T(z)$ is the sufficient statistic, $\eta$ is the canonical parameter, and $A(\eta)$ is the cumulant.
For the von Mises-Fisher distribution, which has the form:
\begin{align}
    C_n(\kappa)\exp(\kappa \mu^T z)
\end{align}
we have $h(z)=1$, $T(z)=z$ and $A(\eta)$ is the negative log of the normalizing constant $C_n(\kappa)$. Instead of a general parameter vector $\eta$, the standard von Mises-Fisher distribution uses a unit vector $\mu=\eta/||\eta||$ and a scale or concentration parameter $\kappa=||\eta||$. 

If $e(z|x)$ is parameterized by a deterministic neural network for the von Mises-Fisher canonical parameter denoted $\overline{e}(x)$, then we have:
\begin{align}
    e(z|x)=C_n(||\overline{e}(x)||)\exp(\overline{e}(x)^T z)
\end{align}
and $\operatorname{KL}[ e(z|x) || e(z|y) ]$ (define $y = x + \Delta x$) is:
\begin{align}
    \label{eq:vmf_kl}
    (\overline{e}(x)-\overline{e}(y))^T \overline{z}(x) + \log C_n(||\overline{e}(x)||) - \log C_n(||\overline{e}(y)||)
\end{align}
where $\overline{z}(x)$ is the mean direction function of the distribution ($\overline{e}(x) = ||\overline{e}(x)||\overline{z}(x)$).
The symmetric KL-divergence ($\operatorname{KL}[ e(z|x) || e(z|y) ] + \operatorname{KL}[ e(z|y) || e(z|x) ]$) is then:
\begin{align}
    (\overline{e}(x)-\overline{e}(y))^T (\overline{z}(x)-\overline{z}(y))
\end{align}
which is closely related to the $L_2^2$ norm of the vector $\overline{e}(x)-\overline{e}(y)$.

Furthermore, we can choose $\kappa=||\overline{e}(x)||$ as a hyperparameter and just parameterize $e(z|x)$'s unit length mean direction $\overline{z}(x)$.
Apart from choosing different $\kappa$ hyperparameters, this is exactly what we do in the C-SimCLR setting described in \Cref{sec:simclr}.

Specifically, in \Cref{sec:simclr}, minimizing the residual information term $I(X;Z|Y)$ correspond to minimizing $\operatorname{KL}[ e(z|x) || b(z|y) ]$ instead of $\operatorname{KL}[ e(z|x) || e(z|y) ]$, where $b$ and $e$ have the same mean direction parameterization but different $\kappa$ hyperparameters, say $\kappa_e$ and $\kappa_b$. 
We can show that the two $\operatorname{KL}$s are actually consistent as learning objectives.
With $\kappa_e, \kappa_b$ as hyperparameters, $\operatorname{KL}[ e(z|x) || e(z|y) ]$ (\Cref{eq:vmf_kl}) can be written as
\begin{align}
    \kappa_e(\overline{z}(x)-\overline{z}(y))^T \overline{z}(x) + \log C_n(\kappa_e) - \log C_n(\kappa_e),
\end{align}
and $\operatorname{KL}[ e(z|x) || b(z|y) ]$ can be written as
\begin{align}
    (\kappa_e\overline{z}(x)-\kappa_b\overline{z}(y))^T \overline{z}(x) + \log C_n(\kappa_e) - \log C_n(\kappa_b) \\
    = (\kappa_e - \kappa_b) +  \kappa_b(\overline{z}(x)-\overline{z}(y))^T \overline{z}(x) + \log C_n(\kappa_e) - \log C_n(\kappa_b).
\end{align}
It is not difficult to see that the two $\operatorname{KL}$s are only different in scale and by a constant, and thus are consistent as learning objectives.
As we claimed in \Cref{sec:lipschitz} (after \Cref{eq:bidir_residual1}), the use of different constant hyperparameters $\kappa_e$ and $\kappa_b$ in the encoders of $x$ and $y$ only changes the minimum achievable $\operatorname{KL}$ divergences.
We can reach the same conclusion for the residual information in another direction $I(Y;Z|X)$.
Thus, whether or not $\kappa_e$ and $\kappa_b$ are the same, we are still minimizing the Lipschitz constant of our encoder function at each observed $(x,y)$ pair when we minimize the residual information terms in the bidirectional CEB objective (\Cref{eq:ceb_bidir}).

\begin{figure}[t]
    \centering
    \includegraphics[keepaspectratio, width=1.0\textwidth]{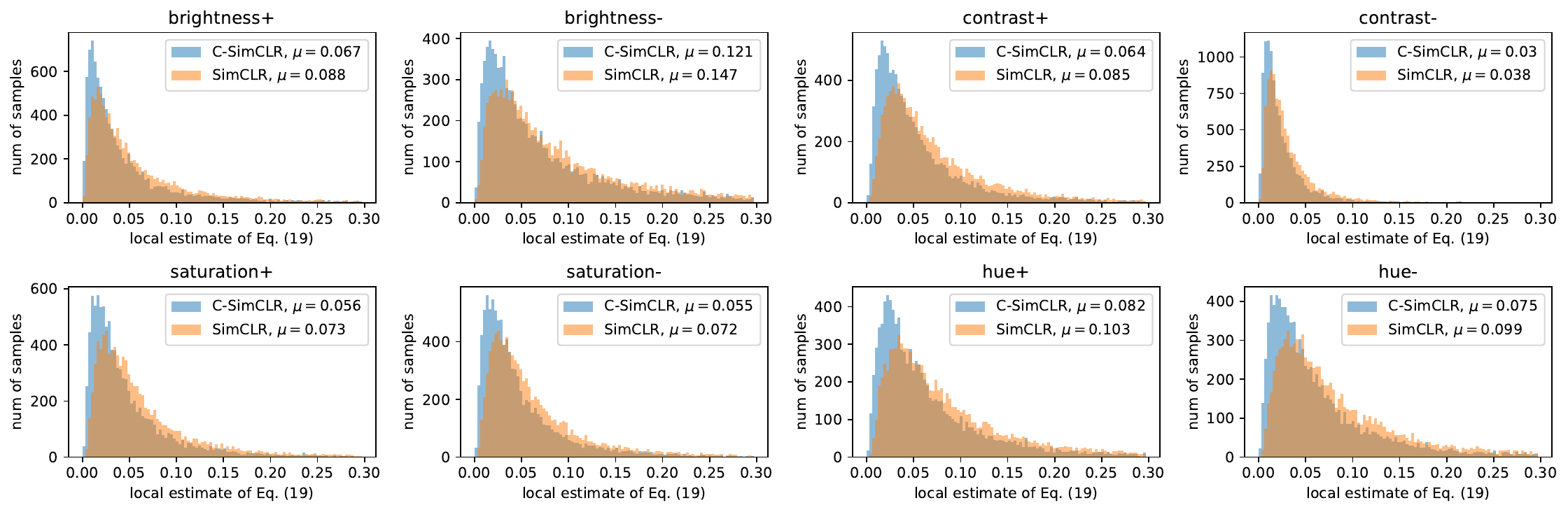}
    \caption{%
      Histograms of \Cref{eq:encoder_lipschitz_main} (also \Cref{eq:encoder_lipschitz} in this section) on 10,000 training images.
      Each local estimate is \Cref{eq:encoder_lipschitz} with a $(x, x + \Delta x)$ pair. 
      Here $x$ is the original image and $x + \Delta x$ is the augmented image.
      SimCLR is in orange.
      C-SimCLR is in blue.
      Higher $y$-axis values for lower $x$-axis values are better.
      We also report the mean ($\mu$) values.
      C-SimCLR consistently outperforms SimCLR.
    }
    \label{fig:lipschitz_train}
\end{figure}

\begin{figure}[t]
    \centering
    \includegraphics[keepaspectratio, width=1.0\textwidth]{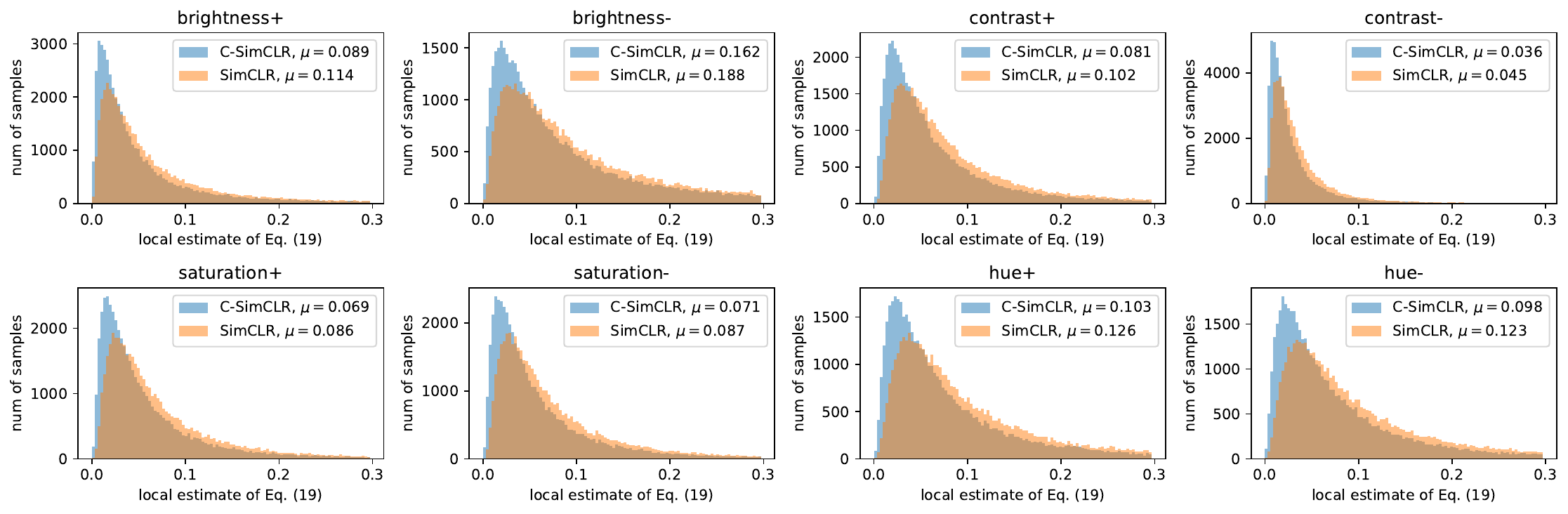}
    \caption{%
      The same as \Cref{fig:lipschitz_train}, but on 50,000 validation images.
    }
    \label{fig:lipschitz_val}
\end{figure}

\paragraph{Estimating the local Lipschitz constant.}
We can evaluate \Cref{eq:encoder_lipschitz_main} (also \Cref{eq:encoder_lipschitz} on any $(x,x + \Delta x)$ pairs to estimate how smooth our model is at that point, and to compare the relative smoothness of different models.
Here, we consider $(x,x + \Delta x)$ pairs where $x$ is taken either from the training or the validation set (using only a center crop in both cases), and $x + \Delta x$ is generated by either increasing or decreasing exactly one of: brightness, contrast, saturation, or hue.
The absolute changes are the maximum adjustment strength in our image defined in \Cref{tab:image_aug} (e.g. for increasing brightness, we increase by 0.4).
In \Cref{fig:lipschitz_train,fig:lipschitz_val}, we compare the histograms of \Cref{eq:encoder_lipschitz_main} on the SimCLR ResNet-50 model and the corresponding C-SimCLR ResNet-50 model.
On both datasets and all eight augmentations, the C-SimCLR models have substantially more mass in the lower values of the local Lipschitz estimates for those image pairs, and have lower mean values computed over the dataset.
Additionally, the mean C-SimCLR results on the \emph{validation} set are almost all lower or equal to the mean SimCLR results on the \emph{training} set, so the smoothness improvements from adding compression to SimCLR appear to be substantial.
The only exceptions are for `brightness+' (SimCLR training mean: 0.088, C-SimCLR validation mean: 0.089) and `brightness-' (SimCLR training mean: 0.147, C-SimCLR validation mean: 0.162).

\section{Pseudocode}
\label{sec:pseudocode}

Listings~\ref{code:csimclr} and~\ref{code:cbyol} show Tensorflow pseudocode for C-SimCLR and C-BYOL respectively.

\begin{listing}[ht]
\begin{minted}[fontsize=\small]{python}
tfd = tensorflow_probability.distributions

def simclr_ceb_loss(x,
                    y,
                    f_enc,
                    kappa_e=1024.0,
                    kappa_b=10.0,
                    beta=1.0):
  """Compute a Contrastive version of CEB loss for C-SimCLR model.

  In practice, we follow SimCLR to apply this loss in a bidirectional manner as
  loss = simclr_ceb_loss(x, y) + simclr_ceb_loss(y, x).
  We use the same notation as the main paper.

  Args:
    x: An augmented image view. The expected shape is [B, H, W, C].
    y: An augmented image view. The expected shape is [B, H, W, C].
    f_enc: An image encoder (Conv + Projection in Fig. 1).
    kappa_e: A float. Concentration parameter of distribution e.
    kappa_b: A float. Concentration parameter of distribution b.
    beta: CEB beta for controlling compression strength (Equation 1).

  Returns:
    A tensor `loss`. The loss is per-sample.
  """
  # Obtain unit-length mean direction vectors with expected shape [B, r_dim].
  r_x = tf.math.l2_normalize(f_enc(x), -1)
  r_y = tf.math.l2_normalize(f_enc(y), -1)

  batch_size = tf.shape(r_x)[0]
  labels_idx = tf.range(batch_size)
  # Labels are pseudo-labels which mark corresponding positives in a batch
  labels = tf.one_hot(labels_idx, batch_size)
  mi_upper_bound = tf.math.log(tf.cast(batch_size, tf.float32))

  e_zx = tfd.VonMisesFisher(r_x, kappa_e)
  b_zy = tfd.VonMisesFisher(r_y, kappa_b)
  z = e_zx.sample()
  log_e_zx = e_zx.log_prob(z)
  log_b_zy = b_zy.log_prob(z)
  i_xzy = log_e_zx - log_b_zy  # residual information I(X;Z|Y)
  logits_ab = b_zy.log_prob(z[:, None, :])  # broadcast

  # The following categorical corresponds to c(y|z) and d(x|z) in Equation 12:
  cat_dist_ab = tfd.Categorical(logits=logits_ab)
  h_yz = -cat_dist_ab.log_prob(labels_idx)
  i_yz = mi_upper_bound - h_yz
  loss = beta * i_xzy - i_yz

  return loss
\end{minted}
\caption{Tensorflow pseudocode of C-SimCLR.}
\label{code:csimclr}
\end{listing}

\begin{listing}

\begin{minted}[fontsize=\small]{python}
tfd = tensorflow_probability.distributions

def byol_ceb_loss(x,
                  x_prime,
                  f_enc,
                  f_enc_target,
                  q_net,
                  l_net,
                  m_net,
                  kappa_e=16384.0,
                  kappa_b=10.0,
                  beta=1.0,
                  byol_loss_weight=2.0):
  """Compute loss for C-BYOL model.

  The notation corresponds to Section 2.3 and Figure 2 of the paper.
  This code presents an updated version of C-BYOL as described in the
  general response.
  
  In practice, we follow BYOL to apply this loss in a bidirectional manner as
  loss = byol_ceb_loss(x, x_prime, ...) + byol_ceb_loss(x_prime, x, ...).
  We use the same notation as the main paper.

  Args:
    x: An augmented image view. The expected shape is [B, H, W, C].
    x_prime: An augmented image view. The expected shape is [B, H, W, C].
    f_enc: An image encoder (Conv + Projection in Fig. 2).
    f_enc_target: The target image encoder. A slow moving-average of f_enc.
    q_net: The BYOL predictor, which is a two-layer MLP.
    l_net: A transformation function. We choose a linear layer in this work.
    m_net: A transformation function. We choose a two-layer MLP in this work.
    kappa_e: A float. Concentration parameter of distribution e.
    kappa_b: A float. Concentration parameter of distribution b.
    beta: CEB beta for controlling compression strength (Equation 1).
    byol_loss_weight: BYOL loss weight. byol_loss_weight = kappa_d/2.

  Returns:
    A tensor `loss`. The loss is per-sample.
  """
  r = f_enc(x)
  mu_e = tf.math.l2_normalize(q_net(r), -1)
  e_zx = tfd.VonMisesFisher(mu_e, kappa_e)
  z = e_zx.sample()  
  y_pred = tf.math.l2_normalize(l_net(z), -1)
  
  r_t = tf.math.l2_normalize(f_enc_target(x), -1)
  y = tf.stop_gradient(r_t)
  mu_b = tf.math.l2_normalize(m_net(y), -1)
  b_zy = tfd.VonMisesFisher(mu_b, kappa_b)
  
  r_t_prime = tf.math.l2_normalize(f_enc_target(x_prime), -1)
  y_prime = tf.stop_gradient(r_t_prime)

  # byol_loss corresponds to -log d(y|z) as described in Section 2.3
  byol_loss = tf.reduce_sum(tf.math.square(y_pred - y_prime), axis=-1)

  log_e_zx = e_zx.log_prob(z)
  log_b_zy = b_zy.log_prob(z)
  i_xzy = log_e_zx - log_b_zy

  loss = byol_loss_weight * byol_loss + beta * i_xzy

  return loss
\end{minted}
\caption{Tensorflow pseudocode of C-BYOL.}
\label{code:cbyol}
\end{listing}